\documentclass[10pt,twocolumn,letterpaper]{article}

\usepackage{cvpr}
\usepackage[utf8]{inputenc} 
\usepackage[T1]{fontenc}    
\usepackage{lscape}
\usepackage{graphicx}
\usepackage{amsmath}
\usepackage{amssymb}
\usepackage{booktabs}
\usepackage{enumitem}
\usepackage{algorithm}
\usepackage{algorithmic}
\usepackage{float} 
\usepackage{acronym}
\usepackage{balance}
\usepackage{xspace}
\usepackage{setspace}
\usepackage{tabularx,colortbl,multirow,array,makecell}
\usepackage{overpic,wrapfig}
\usepackage{nicefrac}
\usepackage[misc]{ifsym}
\usepackage{siunitx}
\usepackage{pifont}
\definecolor{cvprblue}{rgb}{0.21,0.49,0.74}
\usepackage[pagebackref,breaklinks,colorlinks,allcolors=cvprblue]{hyperref}
\usepackage[capitalize]{cleveref}
\usepackage{flushend}

\definecolor{gblue}{HTML}{4285F4}
\definecolor{gred}{HTML}{DB4437}
\definecolor{ggreen}{HTML}{0F9D58}
\definecolor{vblue}{HTML}{2993ba}

\definecolor{gbest}{HTML}{FFCCCB}
\definecolor{gsecond}{HTML}{FFE5CC}
\definecolor{gthird}{HTML}{FFF2A0}
\definecolor{gleast}{HTML}{CCE5FF}

\definecolor{gfinetune}{RGB}{61,108,193}
\definecolor{gzeroshot}{RGB}{229,130,66}
\definecolor{gexist}{RGB}{126,171,85}

\newcommand{\dataset}{\mbox{SceneVerse++}\xspace}
\newcommand{\sdataset}{\mbox{SV++}\xspace}
\newcommand{\datasize}{\mbox{6,687}\xspace}

\newcommand{\rawvidsize}{\mbox{8,217}\xspace}
\newcommand{\supp}{\textit{supplementary}\xspace}

\acrodef{e2e}[E2E]{end-to-end}
\acrodef{vlm}[VLM]{Vision-Language Model}
\acrodef{sfm}[SfM]{Structure-from-Motion}
\acrodef{nvs}[NVS]{novel-view synthesis}
\acrodef{qa}[QA]{question-answering}
\acrodef{cot}[CoT]{Chain-of-Thought}
\acrodef{mllm}[MLLM]{Multimodal Large Language Model}
\acrodef{mca}[MCA]{Multiple-Choice Answers}
\acrodef{na}[NA]{Numerical Answers}

\acrodef{vac}[VAC]{Video Amodal Completion}
\acrodef{vas}[VAS]{Video Amodal Segmentation}
\acrodef{ovo}[OvO]{Object-video-Overlay}
\acrodef{vln}[VLN]{Vision-Lanugage Navigation}
\acrodef{vqa}[VQA]{Visual Question Answering}
\acrodef{scannet}[SN]{ScanNet}
\acrodef{scannetpp}[SN++]{ScanNet++}
\acrodef{tsdf}[TSDF]{Truncated Signed Distance Function}

\makeatletter
\renewcommand{\paragraph}{%
  \@startsection{paragraph}{4}{\z@}%
  {1ex plus 0.5ex minus 0.2ex} 
  {-1em}                      
  {\normalfont\normalsize\bfseries} 
}
\makeatother

\def\paperID{} 
\def\confName{CVPR}
\def\confYear{2026}

\title{Lifting Unlabeled Internet-level Data for 3D Scene Understanding}

\author{
  Yixin Chen\textsuperscript{1}\quad
  Yaowei Zhang\textsuperscript{1}\quad 
  Huangyue Yu\textsuperscript{1}\quad
  Junchao He\textsuperscript{1,2}\quad
  Yan Wang\textsuperscript{1}\\ 
  Jiangyong Huang\textsuperscript{1,3}\quad
  Hongyu Shen\textsuperscript{1,4}\quad
  Junfeng Ni\textsuperscript{1,5}\quad 
  Shaofei Wang\textsuperscript{1}\\
  Baoxiong Jia\textsuperscript{1}\quad
  Song-Chun Zhu\textsuperscript{1,3,5}\quad
  Siyuan Huang\textsuperscript{1}
  \vspace{3pt}\\
  \small \textsuperscript{1} State Key Laboratory of General Artificial Intelligence, BIGAI \quad
  \textsuperscript{2} Beijing University of Posts and Telecommunications \\
  \small \textsuperscript{3} Peking University \quad
  \textsuperscript{4} Beijing Institute of Technology \quad
  \textsuperscript{5} Tsinghua University \quad
  \vspace{3pt}\\
    \href{https://sv-pp.github.io/}{https://sv-pp.github.io/}
    \vspace{-5pt}
}

\begin{document}

\twocolumn[{
\renewcommand\twocolumn[1][]{#1}
\maketitle
\vspace{-0.4in}
\begin{center}
    \centering
    \captionsetup{type=figure}
        \includegraphics[width=\linewidth]{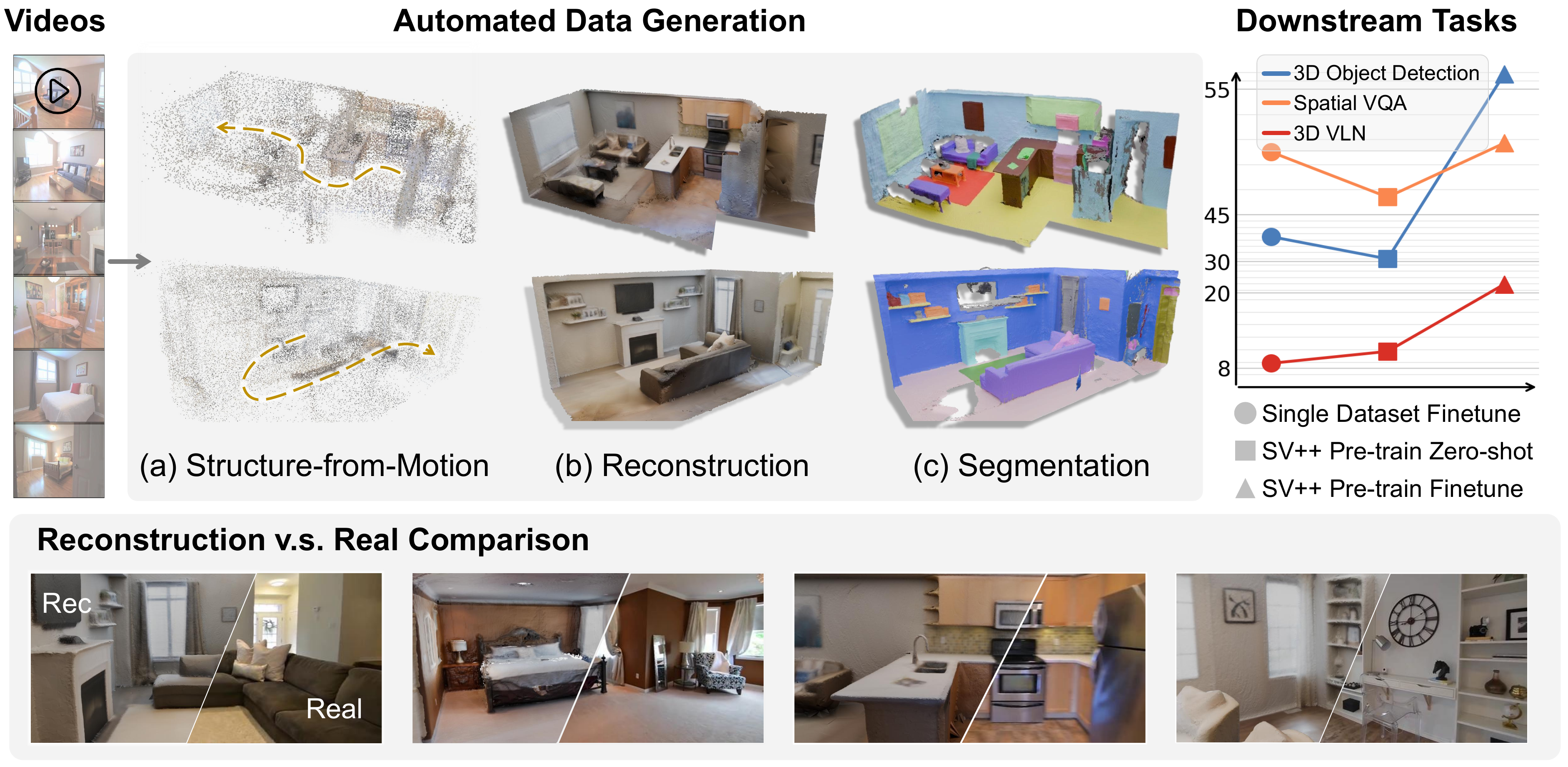}
    \captionof{figure}{\textbf{Overview of \dataset.} From unlabeled internet videos, we build automated data engines to create training data for comprehensive 3D scene understanding, realizing strong zero-shot performance on {existing benchmarks}, with further improvement after {finetuning}. This pinpoints future direction towards 3D spatial intelligence through improved automation on unlabeled, web-scale data.
    }
    \label{fig:teaser}
\end{center}
}]

\begin{abstract}
Annotated 3D scene data is scarce and expensive to acquire, while abundant unlabeled videos are readily available on the internet. In this paper, we demonstrate that carefully designed data engines can leverage web-curated, unlabeled videos to automatically generate training data, to facilitate end-to-end models in 3D scene understanding alongside human-annotated datasets.
We identify and analyze bottlenecks in automated data generation, revealing critical factors that determine the efficiency and effectiveness of learning from unlabeled data.
To validate our approach across different perception granularities, we evaluate on three tasks spanning low-level perception, \ie, 3D object detection and instance segmentation, to high-level reasoning, \ie, 3D spatial \acf{vqa} and \acf{vln}.
Models trained on our generated data demonstrate strong zero-shot performance and show further improvement after finetuning. This demonstrates the viability of leveraging readily available web data as a path toward more capable scene understanding systems. 

\end{abstract}

\section{Introduction}
\label{sec:intro}
With the crucial role of 3D scene understanding in human and embodied intelligence, the field has made remarkable strides in recent years, spanning tasks from geometric perception (\eg, depth estimation~\cite{eigen2014depth,fu2018deep,deng2022depth,esser2023structure,chung2024depth}, camera pose estimation~\cite{hartley2003multiple,schonberger2016structure,wang2024dust3r,wang2024vggsfm,wang2025vggt}), semantic understanding (\eg, 3D object detection~\cite{ding2019votenet,misra2021end,kolodiazhnyi2025unidet3d} and segmentation~\cite{schult2023mask3d,takmaz2023openmask3d,jiang2020pointgroup}) to high-level reasoning (\eg, 3D visual grounding~\cite{chen2020scanrefer,achlioptas2020referit3d,zhu20233d} and spatial reasoning~\cite{azuma2022scanqa,ma2022sqa3d,anderson2018vln,yang2025thinking}). The success of deep learning in this domain is fundamentally tied to the availability of large-scale, annotated, real-world 3D datasets~\cite{dai2017scannet,yeshwanth2023scannet++,mao2022multiscan,baruch2021arkitscenes}.

While methods~\cite{dust3r_cvpr24,wang2025vggt,SpatialLM} in 3D scene understanding continue to improve, progress in 3D scene data with high-quality annotations, on the contrary, has largely stagnated. 
Unlike 2D images~\cite{changpinyo2021conceptual,schuhmann2022laion}, which can be easily scraped and annotated from the web, capturing and labeling 3D data is far more challenging. The common procedure for 3D scene data curation involves recording thousands of frames with specialized hardware, \eg, RGB-D sensors or LiDAR, reconstructing 3D meshes, and manually labeling 3D structures for dense semantic annotations.
In fact, academia has not seen a quantitative leap in 3D data scaling since the pioneering ScanNet~\cite{dai2017scannet}; instead, efforts have focused on simplifying the procedure to get more scenes, \eg, ARKitScenes~\cite{baruch2021arkitscenes} with 2x real-world sites at the cost of coarser scans and labeling, or improving data quality on a manageable number of scans, \eg, ScanNet++~\cite{yeshwanth2023scannet++}.

In this paper, we show that leveraging carefully designed data engines to generate training data from unlabeled, web-scale videos is a promising approach to address the scarcity of annotated 3D scenes. These data engines, often modularized, draw upon prior knowledge from existing foundation models~\cite{li2023blip2,radford2021learning,bai2025qwen2} or scene-specific optimization methods that target particular aspects of general scene understanding, \eg, reconstruction~\cite{mildenhall2020nerf,kerbl3Dgaussians,ni2024phyrecon,ni2026g4splat}, instance segmentation~\cite{yan2024maskclustering,fu2022panoptic,kundu2022panoptic}, and open-set semantics~\cite{peng2023openscene,liu2023weakly,justin2023lerf}. Since these submodular methods vary in representation, methodology, and technical focus, design choices for automatic data generation are non-trivial. The effectiveness of scaling generated data is task-dependent and strongly influenced by both quality and efficiency considerations. 

To this end, we systematically analyze the bottlenecks in creating automated data engines for 3D scene understanding, provide guidelines on how to scale \ac{e2e} models, and pinpoint what submodular models should prioritize in future development.
From Internet videos, we curate \dataset of \datasize real-world scenes with images, camera poses, dense reconstructions, instance segmentations and high-level reasoning annotations. We demonstrate the effectiveness of internet-scale data by empowering three exemplar tasks in 3D scene understanding:
\begin{itemize}
    \item 3D detection and segmentation. The models trained on \dataset realize strong zero-shot performance on ScanNet and ARKitscenes, and further significantly improve after finetuning (+20.6 for F1@.25).
    \item 3D spatial \acf{vqa}: Training on \dataset significantly improves the spatial reasoning performance of \acp{vlm}, achieving zero-shot performance comparable to models trained on ground-truth 3D scenes.
    \item 3D \acf{vln}: We examine the zero-shot transfer from real-world videos to navigation in simulation, and demonstrate \dataset brings an extra 14\% navigation success rate after finetuning.
\end{itemize}

\section{Related Work}

\subsection{3D Scene Understanding and Datasets}

Early work in 3D scene understanding primarily focuses on tasks such as semantic segmentation~\cite{qi2017pointnet++,wang2019dgcnn,zhang2022pointclip}, instance segmentation~\cite{schult2023mask3d,takmaz2023openmask3d,jiang2020pointgroup,zhu2024pq3d}, and object detection from images~\cite{chen2016monocular,ding2019votenet,brazil2023omni3d} or point clouds~\cite{misra2021end,SpatialLM,kolodiazhnyi2025unidet3d,avetisyan2024scenescript}. Beyond geometry-centric perception, there has been growing interest in vision-language tasks within 3D scenes, including object referral~\cite{chen2020scanrefer,achlioptas2020referit3d,zhang2023multi3drefer}, captioning~\cite{chen2021scan2cap,yuan2022x,chen2021d3net,chen2023end}, spatial reasoning~\cite{azuma2022scanqa,ma2022sqa3d,hong20233d,yang2025thinking,chen2025synergai}, and navigation~\cite{hong2021vln,anderson2018vln,ku2020room,qi2020reverie}.
The shift is driven by the popularity of \ac{e2e} \acp{vlm}~\cite{openai2023gpt4,team2023gemini,bai2025qwen2,chen2024internvl}, offering advantages in multi-tasking~\cite{kojima2022large} and scaling~\cite{kaplan2020scaling,jia2024sceneverse,shukor2025scaling} in both model architecture and training data. 

The success of these \ac{e2e} models relies critically on 3D datasets~\cite{mao2022multiscan,wald2019rio,ramakrishnan2021habitat,khanna2024habitat,zheng2020structured3d,yu2025metascenes} with detailed annotation, 
such as pioneering ScanNet~\cite{dai2017scannet}, later ARKitScenes~\cite{baruch2021arkitscenes} captured with portable devices, and ScanNet++~\cite{yeshwanth2023scannet++} with higher-quality scans. However, unlike their 2D counterparts, the scaling of the 3D datasets faces significant bottlenecks in capture and labeling costs that hinder further expansion. In the meantime, the internet contains orders of magnitude more unlabeled data that captures our 3D world.

In this paper, we advocate for advancing comprehensive 3D scene understanding by leveraging these unlabeled internet videos. We build upon methods that address intermediate problems in scene understanding, achieved by leveraging pre-trained models in a training-free~\cite{curless1996volumetric,dust3r_cvpr24,luo2023scalable,yan2024maskclustering} or weakly-supervised manner~\cite{liu20243dgs,fu2022panoptic,shen2025trace3d,shafiullah2022clipfields} to inject knowledge into 3D, \eg, open-vocabulary 3D segmentation by lifting 2D results~\cite{takmaz2023openmask3d}. We build automated data engines on top of these submodules, leveraging their complementary strengths while mitigating limitations, achieving an efficiency-efficacy balance in internet-level data scaling.

\subsection{Leveraging Internet-level Videos}

Recognizing the scarcity of 3D datasets, an emerging direction is to harness video data to lift 2D content into 3D annotations for training. For instance, \citet{miao2025towards} proposes using existing 2D single-view datasets with estimated depth to generate 3D annotations. However, their data generation is bound to existing datasets~\cite{lin2014microsoft,shao2019objects365} with 2D segmentation annotations and operates at the single-image level, presenting a significant gap towards whole-scene understanding. The abundant internet videos present an attractive, untapped resource, and recent work has begun to explore this direction, but mostly on training generative video~\cite{Ma2025See3D,ali2025cosmos,wan2025,ho2022imagen,lu2025taco} or \ac{nvs}~\cite{liu2024novel,sargent2024zeronvs,dong2023ivs,lu2024movis} models. In pursuit of scalable 3D scene understanding~\cite{lin2023learning,yang2025como,shen2024gim}, RoomTour3D~\cite{han2025roomtour3d} generates video instructions for navigation through summarization and candidate view selection, while NaVILA~\cite{cheng2024navila} incorporates real video trajectories into training to improve instruction-following in \acf{vln}. However, they remain confined to the navigation domain, without addressing broader spatial reasoning or scene understanding. Moreover, they often treat the multi-module data generation pipeline as given, offering little analysis of which components are most critical or where errors propagate. In contrast, our work addresses comprehensive 3D scene understanding tasks, from low-level perception to high-level reasoning, and provides systematic analyses, examining both the efficiency and efficacy of transforming internet-scale data for task-specific training.

\section{Data Curation for \texorpdfstring{\dataset}{}}
\label{sec:curation}
Our work focuses on 3D scene understanding for static indoor scenes. The first step for task-specific 3D scene understanding is to curate internet videos and convert them to a basic 3D representation consisting of camera poses and sparse 3D geometry. Inspired by prior work on internet data processing~\cite{ali2025cosmos,han2025roomtour3d,lin2023learning}, our data pipeline combines video curation with \ac{sfm}~\cite{hartley2003multiple}, encompassing shot splitting, filtering, key frame extraction, pixel matching, global bundle adjustment, and quality check. 

We use TransNetV2~\cite{soucek2020transnetv2} to detect shots in long-form videos and discard very short clips. 
The filtering process removes low-quality or unsuitable content, including pure black screen, visual noise, humans~\cite{he2017mask}, and outdoor scenes~\cite{zhou2017places}. 
To handle potentially long-duration internet videos, we select keyframes based on parallax rather than uniform sampling~\cite{han2025roomtour3d}, ensuring well-constrained triangulation with redundancy control.
For sparse reconstruction and camera pose estimation, we adopt a dense pixel matching and bundle adjustment approach, which provides more robust camera poses and sparse point clouds than existing feed-forward methods~\cite{dust3r_cvpr24,wang2025vggt}. The overall pipeline resembles Mast3R-SFM~\cite{duisterhof2025mast3r}, and we introduce optimized pseudo-track pixels to improve memory efficiency for long-term videos (\eg, >300 frames) and incorporate relative image similarity to address the false-positive bias in existing pixel matching models~\cite{leroy2024grounding}. Finally, we filter out the scenes with small spatial coverage, relatively empty space, or wrong \ac{sfm} results. This can be achieved by existing \acp{vlm}~\cite{openai2023gpt4,bai2025qwen2,team2023gemini}, but we resort to human annotation (<10 seconds/scene) to ensure data quality for downstream tasks. 

\begin{figure}
    \centering
    \includegraphics[width=1.0\linewidth]{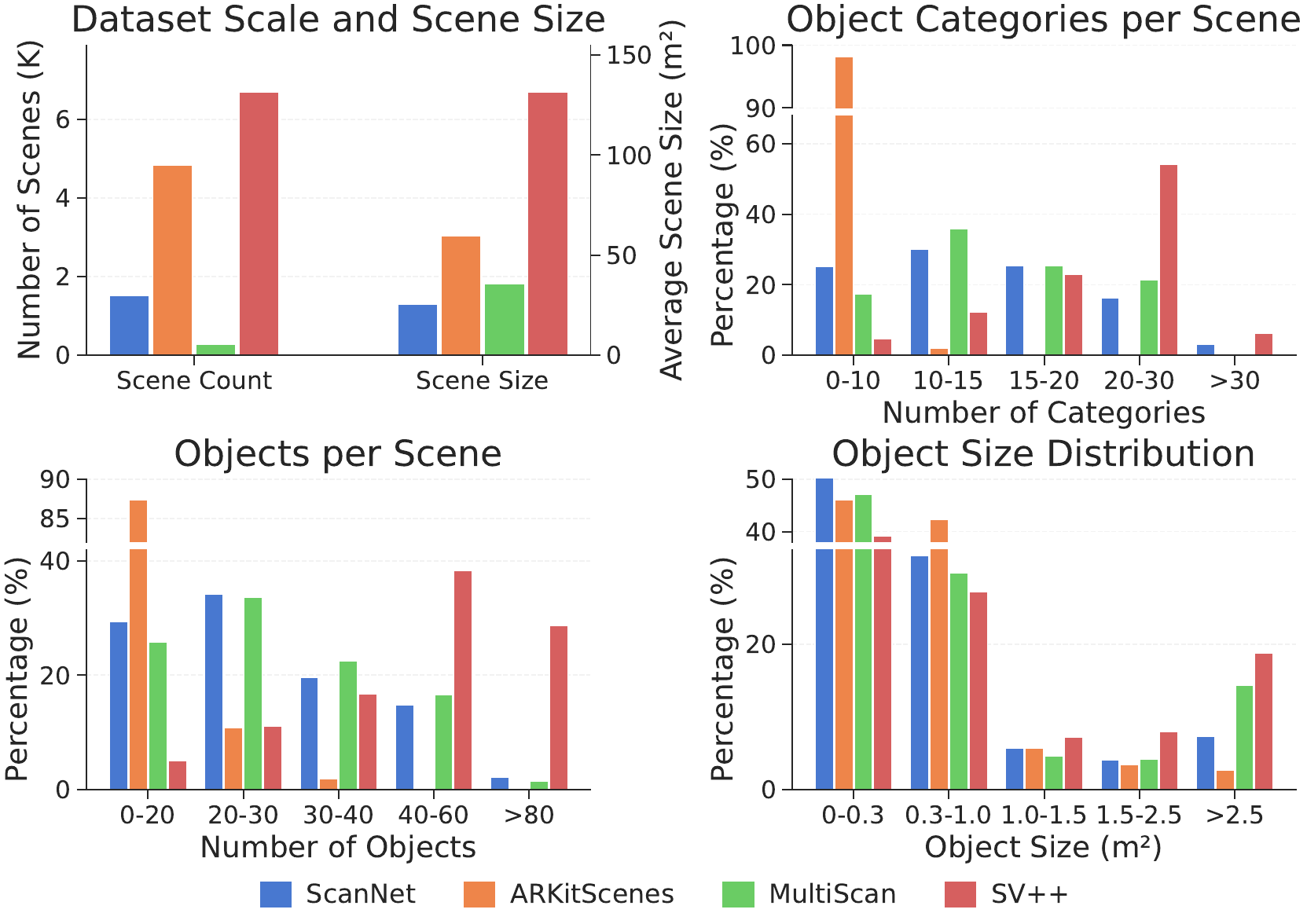}
    \caption{\textbf{Statistics comparison.} \dataset encompasses more scenes, larger areas, and greater object diversity compared with existing real-world datasets.}
    \label{fig:data_stats}
\end{figure}

\paragraph{Statistics}
The dataset statistics are shown in \cref{fig:data_stats}, with comparison with ScanNet~\cite{dai2017scannet}, MultiScan~\cite{mao2022multiscan} and ARKitScenes~\cite{baruch2021arkitscenes}. Starting from \rawvidsize videos from the open internet platforms, we obtain \datasize scenes, exceeding ARKitScenes captured with portable devices. \dataset contains multi-floor, multi-room scans from long-range videos, producing scenes significantly larger\footnote{Scene area is approximated by the product of extents along x-y plane.} than existing room-scale or lab-based datasets. More details about data curation, \ac{sfm} methods, and examples with camera trajectories and sparse geometry are presented in \supp.

\section{\texorpdfstring{\dataset}{} for 3D Scene Understanding}
In this section, we present how to leverage \dataset to generate training data and improve on three representative tasks in 3D scene understanding.
\subsection{3D Object Detection and Segmentation}
\label{sec:segmentation}

\begin{figure*}[!ht]
    \centering
    \includegraphics[width=1.0\linewidth]{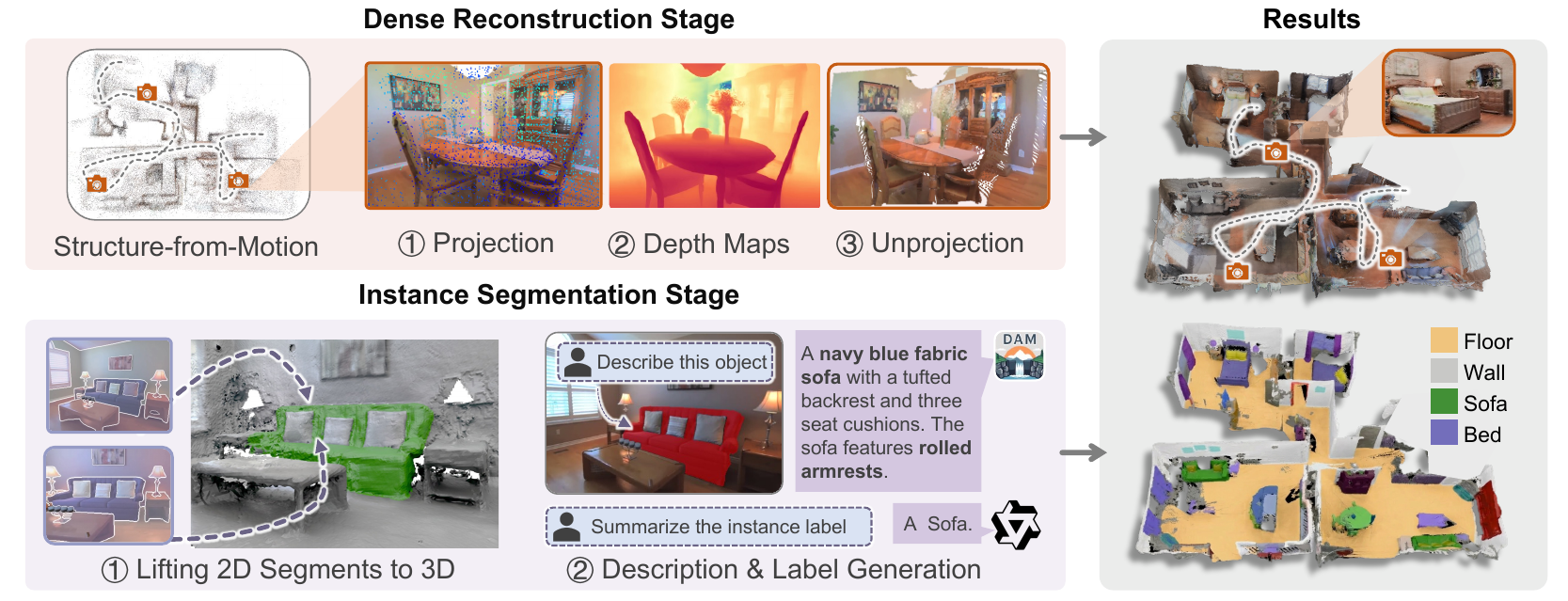}
    \caption{\textbf{Overview of data generation.} The pipeline leverages a modular design for automatic 3D reconstruction and segmentation.}
    \label{fig:segmentation_pipeline}
\end{figure*}

\paragraph{Task and Benchmark}
The 3D object detection and segmentation task aims to localize distinct objects within a 3D scene, assigning each a precise geometric boundary and a semantic label. This task serves as a bridge between low-level 3D reconstruction and high-level scene understanding.
In the following, we first introduce the data engine that generates 3D instance annotations, and then evaluate its effectiveness on real-world benchmarks. 

\paragraph{Data Generation}

To obtain the complete reconstructed meshes and instance-level annotations from the sparse outputs of \ac{sfm}, we design a reconstruction and segmentation pipeline, as illustrated in \cref{fig:segmentation_pipeline}. It transforms internet images into 3D scenes, considering both efficiency and effectiveness in large-scale data generation.

\textit{Dense Reconstruction.}
Recent advances in 3D reconstruction introduce various approaches with different trade-offs between quality and efficiency. Neural rendering methods~\cite{yu2022monosdf,ni2024phyrecon,ni2025dprecon,guedon2025matcha,huang20242dgs,ni2026g4splat,chen2024pgsr} produce photo-realistic rendering and recover detailed geometry, but they require dense computation for per-scene optimization, especially for large and complex environments. \Acl{e2e} reconstruction frameworks~\cite{dust3r_cvpr24,wang2025vggt} enable dense point cloud reconstruction directly from images, providing convenience and speed; however, they struggle with long videos due to memory constraints and often exhibit obvious artifacts in multi-view consistency and geometry distortion. 
    
\begin{figure}
    \centering
    \includegraphics[width=1.0\linewidth]{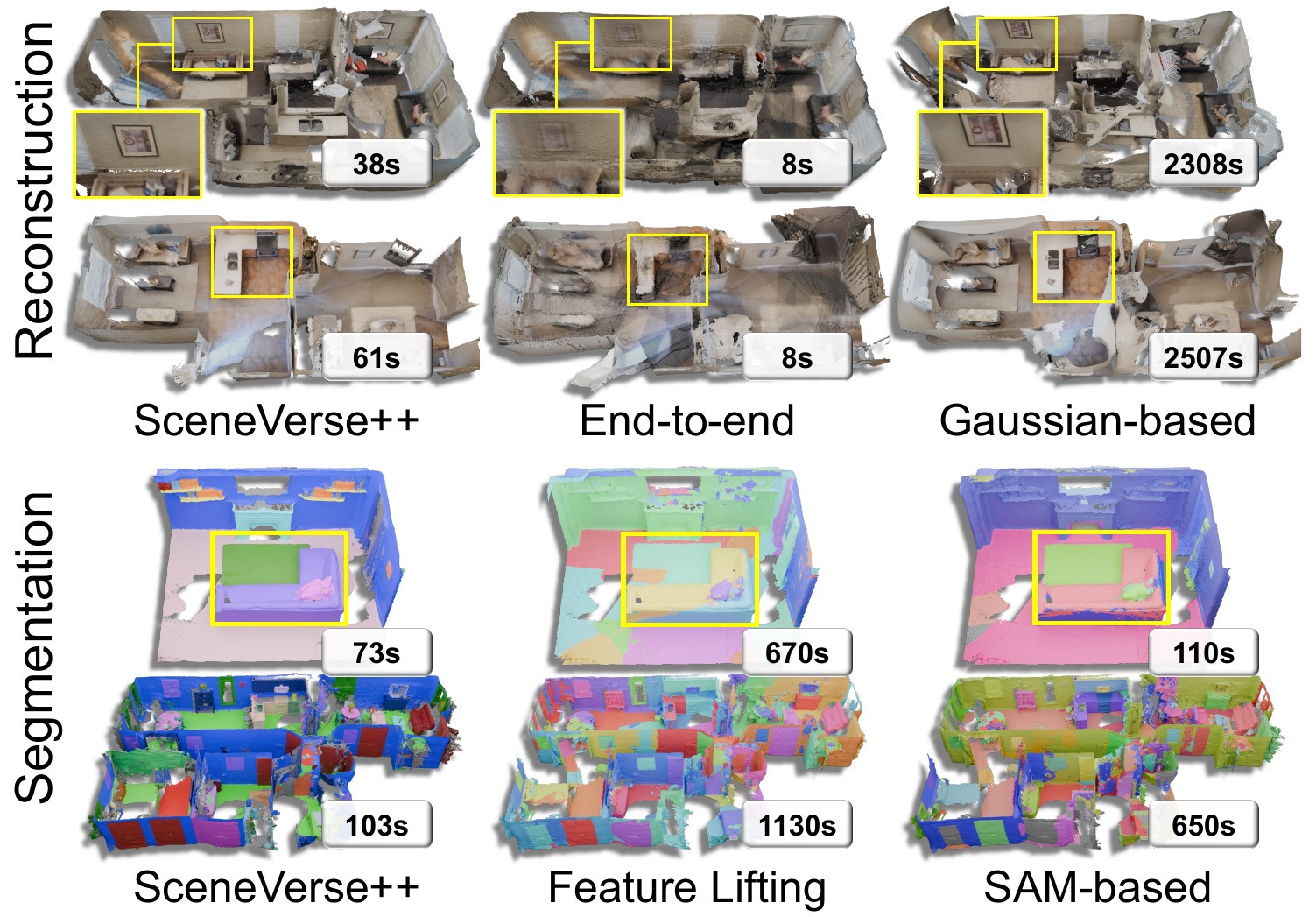}
    \caption{\textbf{Reconstruction and segmentation comparison}, where \dataset features a balance in quality and efficiency.}
    \label{fig:recon_seg}
\end{figure}

To balance efficiency and reconstruction quality, we design a reconstruction pipeline based on metric depth estimation that effectively leverages \ac{sfm} outputs from \cref{sec:curation}. 
Specifically, we project the reconstructed sparse 3D points onto the image plane to obtain sparse depth maps, which serve as priors for PriorDA~\cite{wang2025depthprior} to predict dense metric depth maps.
The predicted depths are then fused using a \acf{tsdf} representation to produce watertight 3D meshes. During fusion, unreliable large depth values are truncated, and radius- and statistical-based filters further remove floating noisy points.
This design achieves stable, high-quality reconstructions with reduced computational cost, enabling efficient processing of large-scale internet videos while maintaining sufficient accuracy for downstream tasks. Qualitative results and computation time comparison are shown in \cref{fig:recon_seg}.

\textit{Instance Segmentation.}
Recent advances in per-scene 3D segmentation have also explored different paradigms. For example, image-based approaches, such as the SAM series~\cite{kirillov2023segment,ravi2024sam2}, effectively identify 2D object masks across frames, but do not explicitly leverage 3D spatial information. When applied to long video sequences, they often produce duplicated instances due to incorrect cross-view associations. In contrast, feature-lifting methods~\cite{fu2022panoptic,bhalgat2023contrastive,shen2025trace3d} exploit spatial correspondences across multiple views through rendering~\cite{mildenhall2020nerf,kerbl3Dgaussians}, but their performance is affected by the rendering quality and typically requires substantial computational resources and processing time for long videos.
    
To overcome these challenges, we choose to lift 2D masks to 3D using the dense reconstruction results. Specifically, we first apply CropFormer~\cite{qi2023high} to obtain per-frame segmentation masks, which are then aggregated in 3D space based on neighboring-frame view consensus~\cite{yan2024maskclustering} and spatial agreement.
Finally, we employ Describe Anything~\cite{describeanything} and Qwen2-VL~\cite{bai2025qwen2} to automatically generate textual descriptions for each 3D instance and align their semantic labels to the ScanNet category set. The segmentation comparison is shown in \cref{fig:recon_seg}.

\paragraph{Statistics}
In practice, the average runtime for each scene is 71 seconds for dense reconstruction and 96 seconds for segmentation. On average, each scene in \dataset has 49 objects across 21 distinct categories, both surpassing existing datasets as shown in \cref{fig:data_stats}. This reflects the greater diversity of object types and richer scene compositions in our data. In addition, the object size distribution in \dataset closely aligns with that of real-world datasets, indicating that our reconstructed scenes preserve realistic scale and spatial relationships.

\paragraph{Performance}
We validate the effectiveness of our dataset on 3D object detection with SpatialLM~\cite{SpatialLM} and 3D instance segmentation with Mask3D~\cite{schult2023mask3d}. The quantitative results are summarized in \cref{tab:3D_instance_segmentation}.
\begin{itemize}
    \item SpatialLM, derived from a \ac{mllm}, generates structured 3D scene descriptions for object detection and is originally trained on a synthetic dataset of over 12,000 indoor scenes. We adopt the same base model and evaluate on two real-world benchmarks. Without fine-tuning, the model trained on \dataset achieves slightly better detection on ScanNet and ARKitScenes than training only on synthetic data. When fine-tuned on ScanNet, the model pretrained on \dataset achieves a substantial improvement, \ie, F1@0.25 of 58.6 \vs 38.0; this shows \dataset better captures real-world distributions and provides a better initialization. Training from scratch on ScanNet fails to converge, as the adapter linking 3D encoder~\cite{wu2025sonata} to \ac{mllm} requires significant pretraining~\cite{SpatialLM,huang2023embodied}.
    \item The results on 3D instance segmentation using Mask3D reveal a different trend: the model pretrained solely on \dataset does not transfer well to ScanNet, but it consistently improves performance across all metrics after finetuning compared with training from scratch. This drop stems from Mask3D's reliance on segment-level masks obtained from a graph-based segmentation~\cite{felzenszwalb2004efficient}, which is highly sensitive to sensor and reconstruction pipelines. This highlights a key factor in model scaling - their susceptibility to domain-specific bias. 
\end{itemize}
    More details, additional experiments and ablations, and further discussions are provided in \supp.

\begin{table}[!t]
\small
\centering
\caption{\textbf{Testing SpatialLM on 3D object detection.} Performance is reported under different pretraining and finetuning configurations with the same model architecture.}
\label{tab:3d_object_detection}
\resizebox{\linewidth}{!}{
\begin{tabular}{c|cc|cc}
\toprule
\textbf{Benchmark} & \textbf{Pretrain} & \textbf{Finetune} & \textbf{F1@.25} & \textbf{F1@0.5} \\
\midrule
\multirow{2}{*}{\makecell{ARKitScenes}} 
 & SpatialLM & - & 35.1 & 21.2 \\ 
 & \dataset  & - & 35.8 & 20.7 \\ 
\midrule
\multirow{5}{*}{ScanNet} 
 & - & ScanNet & 2.9 & 0.7 \\ 
 & SpatialLM & - & 29.0 & 19.7 \\ 
 & \dataset  & - & 30.9 & 21.3 \\ 
 & SpatialLM & ScanNet & 38.0 & 28.7 \\ 
 & \dataset  & ScanNet & \textbf{58.6} & \textbf{45.4} \\ 
\bottomrule
\end{tabular}}
\end{table}

\begin{table}[!t]
\small
\centering
\caption{\textbf{Testing Mask3D on 3D instance segmentation}. It presents reliance on data-specific bias that hurts model scaling.}
\label{tab:3D_instance_segmentation}
\resizebox{\linewidth}{!}{
\begin{tabular}{c|cc|ccc}
\toprule
\textbf{Benchmark} & \textbf{Pretrain} & \textbf{Finetune} & \textbf{AP$_{25}$} & \textbf{AP$_{50}$} & \textbf{AP}\\
\midrule
\multirow{3}{*}{ScanNet} 
 & - & ScanNet & 36.1 & 31.8 & 22.8\\ 
 & \dataset  & - & 15.4 & 13.0 & 8.3\\ 
 & \dataset  & ScanNet & \textbf{38.5} & \textbf{32.9} & \textbf{23.6} \\ 
\bottomrule
\end{tabular}}
\end{table}

\subsection{3D Spatial VQA}
\label{sec:3dvqa}

\begin{table*}[htbp]
\centering
\begin{minipage}[h]{0.715\textwidth}
\centering
\caption{\textbf{Evaluation results on VSI-Bench.} Performance is reported on both the full set and ARKitScenes subset: 1) zero-shot test (-); 2) trained on \dataset (\sdataset); 3) trained on VLM-3R data from ScanNet and ScanNet++ (SN, SN++); and 4) trained on the combination of 2) and 3) (All). The figures of ``SN, SN++'' and ``All'' on the full set indicate in-domain (ID) results, while others are out-of-domain (OOD) results. \dataset is more effective in improving \colorbox{gsecond}{general spatial knowledge} but less in \colorbox{gleast}{domain-specific knowledge}.}
\label{tab:vsibench}
\resizebox{\linewidth}{!}{
\begin{tabular}{c|c|ccccccccc|c}
\toprule
\multirow{2}{*}{\textbf{Model}} & 
\multirow{2}{*}{\makecell{\textbf{Dataset}\\\textbf{Source}}} & 
\multicolumn{9}{c|}{\textbf{VSI-Bench Fullset}$_\text{SN, SN++, ARKit}$} & \textbf{Subset}$_\text{ARKit}$ \\ 
 &  & 
App. Ord. &
\cellcolor{gleast}Abs. Dist. &
\cellcolor{gleast}Obj. Cnt. &
\cellcolor{gsecond}Rel. Dist. &
\cellcolor{gleast}Obj. Size &
\cellcolor{gleast}Room Size &
Route Plan &
\cellcolor{gsecond}Rel. Dir. &
Avg. &
Avg. \\
\midrule
\multirow{4}{*}{\makecell{Qwen2.5\\-VL-3B}} & - & 27.3 & 17.4 & 25.2 & 37.2 & 16.5 & 26.2 & 28.4 & 45.4 & 27.9 & 28.1 \\ 
 & \sdataset                        & 26.1 & 30.2 & 61.8 & 49.3 & 49.8 & 43.9 & 33.6 & 47.8 & 42.8 & 48.0 \\ 
 & \acs{scannet}, \acs{scannetpp}  & 32.4 & 39.6 & 67.4 & 48.9 & 64.0 & 53.8 & 38.7 & 44.9 & 48.7 & 49.0 \\ 
 & All                             & 27.2 & 39.3 & 67.5 & 50.3 & 63.5 & 54.0 & 36.6 & 55.8 & 49.3 & 51.3 \\ 
\midrule

\multirow{4}{*}{\makecell{Qwen2.5\\-VL-7B}} & - & 34.5 & 21.0 & 41.5 & 38.6 & 50.5 & 36.7 & 29.4 & 41.0 & 36.6 & 39.4 \\ 
 & \sdataset                        & 43.4 & 28.9 & 63.8 & 48.9 & 57.0 & 46.4 & 35.1 & 48.0 & 46.4 & 49.1 \\ 
 & \acs{scannet}, \acs{scannetpp}  & 37.7 & 38.8 & 68.3 & 52.8 & 64.8 & 53.0 & 37.1 & 47.3 & 50.0 & 48.8 \\ 
 & All                             & 29.8 & 38.3 & 67.1 & 51.7 & 65.8 & 53.5 & 41.2 & 57.3 & 50.7 & 50.5 \\ 
\bottomrule
\end{tabular}
}

\end{minipage}%
\hfill
\begin{minipage}[h]{0.28\textwidth}
\centering
\includegraphics[width=\linewidth]{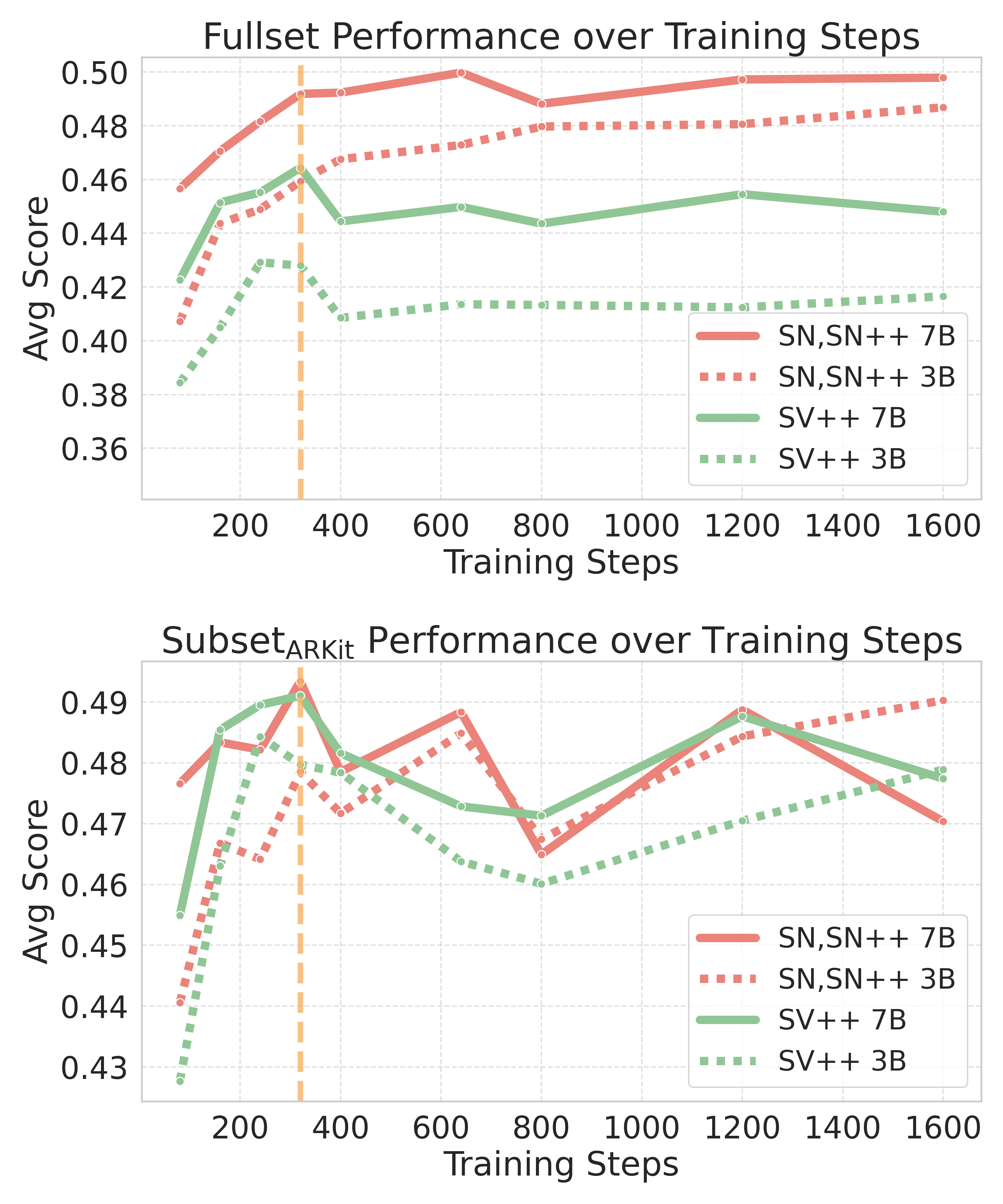}
\vspace{-2.2em}
\captionof{figure}{\textbf{Training dynamics.}}
\label{fig:scaling}
\end{minipage}
\vspace{-1.3em}

\end{table*}

\paragraph{Task and Benchmark}
Visual-spatial intelligence~\cite{yang2025thinking} requires the combination of visual perception, linguistic understanding, temporal reasoning, and spatial reasoning~\cite{herskovits1986language,gardner2011frames}. Despite being a critical capability for future embodied agents to explore and perform tasks in the 3D world, it remains a challenging frontier for current \acp{vlm}.
To investigate how \dataset can improve the spatial reasoning ability of \ac{vlm}, we focus on 3D spatial \acf{vqa}, which requires a model to answer questions about 3D space by inferring spatial relations from 2D visual input. We evaluate on VSI-Bench~\cite{yang2025thinking}, a 3D spatial understanding benchmark constructed from egocentric videos in ScanNet, ScanNet++, and ARKitScenes. It contains over 5,000 \ac{qa} pairs spanning eight task types, presented as Multiple-Choice Answers (MCA) or Numerical Answers (NA). MCA performance is measured by mean accuracy, while NA performance is calculated using relative accuracy across multiple confidence thresholds. 

\paragraph{Data Generation}
We generate general spatial \acp{qa} by transferring the geometry and semantic information in 3D scenes (\cref{sec:segmentation}) to 3D scene graphs~\cite{armeni20193d,wald2020learning,jia2024sceneverse}, following VLM-3R~\cite{fan2025vlm}. Each node in the scene graph represents a distinct 3D object instance, and edges represent pairwise spatial relations. Leveraging these structured semantics, \ac{qa} pairs are automatically generated for \textit{Object Counting, Relative Distance, Relative Direction, Object Size, Absolute Distance}, and \textit{Room Size} by designing task-specific templates~\cite{fan2025vlm}.
For the \textit{Route Planning} task, we generate \ac{qa} pairs by employing a \ac{vlm}~\cite{bai2025qwen2} to summarize the navigation trajectories within 3D environments (introduced in \cref{sec:navigation}). The summary is then transformed into fill-in-the-blank Multiple-Choice questions by masking specific actions. The \textit{Appearance Order} task is not included following the setting of VLM3R.

\paragraph{Statistics}
Applying the automatic generation pipeline to the reconstructed scenes in \dataset yields 632K spatial VQA data following the VSI-Bench format. It comprises 391K samples for MCA and 241K samples for NA, respectively. More details on data generation and question type distribution are in \supp.

\paragraph{Performance}
We evaluate the performance of Qwen2.5-VL after LoRA fine-tuning~\cite{hu2022lora} on VSI-Bench, which spans ScanNet (SN), ScanNet++ (SN++), and ARKitScenes (ARKit). Given the domain discrepancy between datasets, we regard training and testing on SN and SN++ as in-domain (ID), and out-of-domain (OOD) otherwise. 
For fairness, we sample 202K data from \dataset for training, comparable with 206K samples on SN and SN++ from VLM3R~\cite{fan2025vlm}. We report quantitative results in \cref{tab:vsibench} and key observations as follows:

\begin{itemize}
    \item \textbf{Spatial reasoning enhancement.} \dataset can improve the spatial reasoning capability of the base \acp{vlm}, yielding +14.9 for the 3B model and +9.8 for 7B on VSI-Bench full set. This highlights \dataset as a reliable and promising data source for advancing existing \acp{vlm}.
    \item \textbf{Domain generalization.} We observe comparable performance between \dataset and SN/SN++ on the VSI-Bench ARKit subset, indicating their comparable domain generalizability, despite that SN and SN++ have groundtruth annotations. This contrasts with the performance gap observed on the VSI-Bench full set, reflecting a larger-than-expected domain gap across datasets. Training on all data sources (All) further improves performance on both the full set and ARKit subset, showing the benefit of a broader domain covered in \dataset.
    \item \textbf{Category-wise difference.} Per-category analysis reveals that \dataset delivers greater improvement on categories concerning \textit{general spatial knowledge} such as \textit{Relative Distance} and \textit{Relative Direction}, which are less susceptible to domain-specific distribution. In contrast, it exhibits worse results on categories highly relying on \textit{domain-specific knowledge} such as \textit{Object Count} and \textit{Room Size}, likely due to variations in object and scene distributions, as illustrated in \cref{fig:data_stats}.
    \item \textbf{Training dynamics.} We visualize the evolution of evaluation results within one training epoch in \cref{fig:scaling}. A distinct turning point (green dashed line) emerges: model performance consistently improves before this point, after which in-domain training (SN, SN++ curves on full set) continues to rise while others plateau or decline. This provides further evidence of domain gap and overfitting to \textit{domain-specific knowledge}, aligning with findings from concurrent works~\cite{brown2025sims,brown2025benchmark}.
\end{itemize}

\subsection{3D \texorpdfstring{\acf{vln}}{}}
\label{sec:navigation}
\paragraph{Task and Benchmark}
The goal of \ac{vln} is to enable embodied agents to follow natural language instructions and navigate toward specified goals within 3D environments.
Room-tour videos from the internet provide a valuable proxy for natural human navigation in real indoor spaces. Unlike prior work~\cite{lin2023learning,cheng2024navila}, we focus on the key factors to provide rich, continuous trajectories that can bridge the gap between model navigation and real-world embodied behaviour.
We adopt the widely used Room-to-Room (R2R) benchmark~\cite{anderson2018vln} built on Matterport3D~\cite{matterport3d} environments, where an agent receives a sequence of rendered egocentric observations and a goal-directed instruction as input, and outputs a sequence of discrete navigation actions. 
The action space consists of fixed translation and rotation steps, where movements are discretized into three distance bins of [25, 50, 75] cm and rotations into [15°, 30°, 45°]. 

\begin{figure*}[ht]
    \centering
    \includegraphics[width=\textwidth]{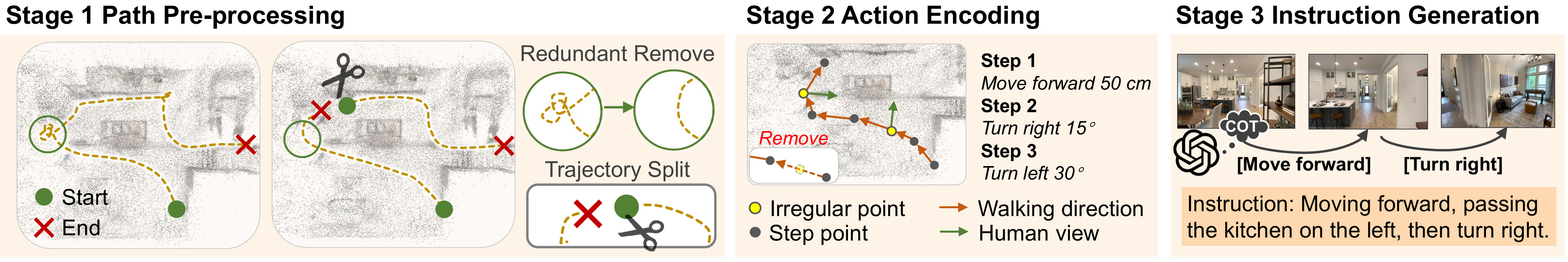}
    \caption{\textbf{Overview of the VLN data generation pipeline.} We construct VLN data from room-tour videos by (i) preprocessing trajectories to eliminate redundant local rotations and segmenting long paths into sub-paths suitable for instruction generation; (ii) converting camera transitions within each sub-path into R2R-style navigation actions; and (iii) generating instructions for each sub-path using VLMs.}
    \label{fig:vln_system}
\end{figure*}

\begin{figure}[ht]
    \centering
    \includegraphics[width=1.0\linewidth]{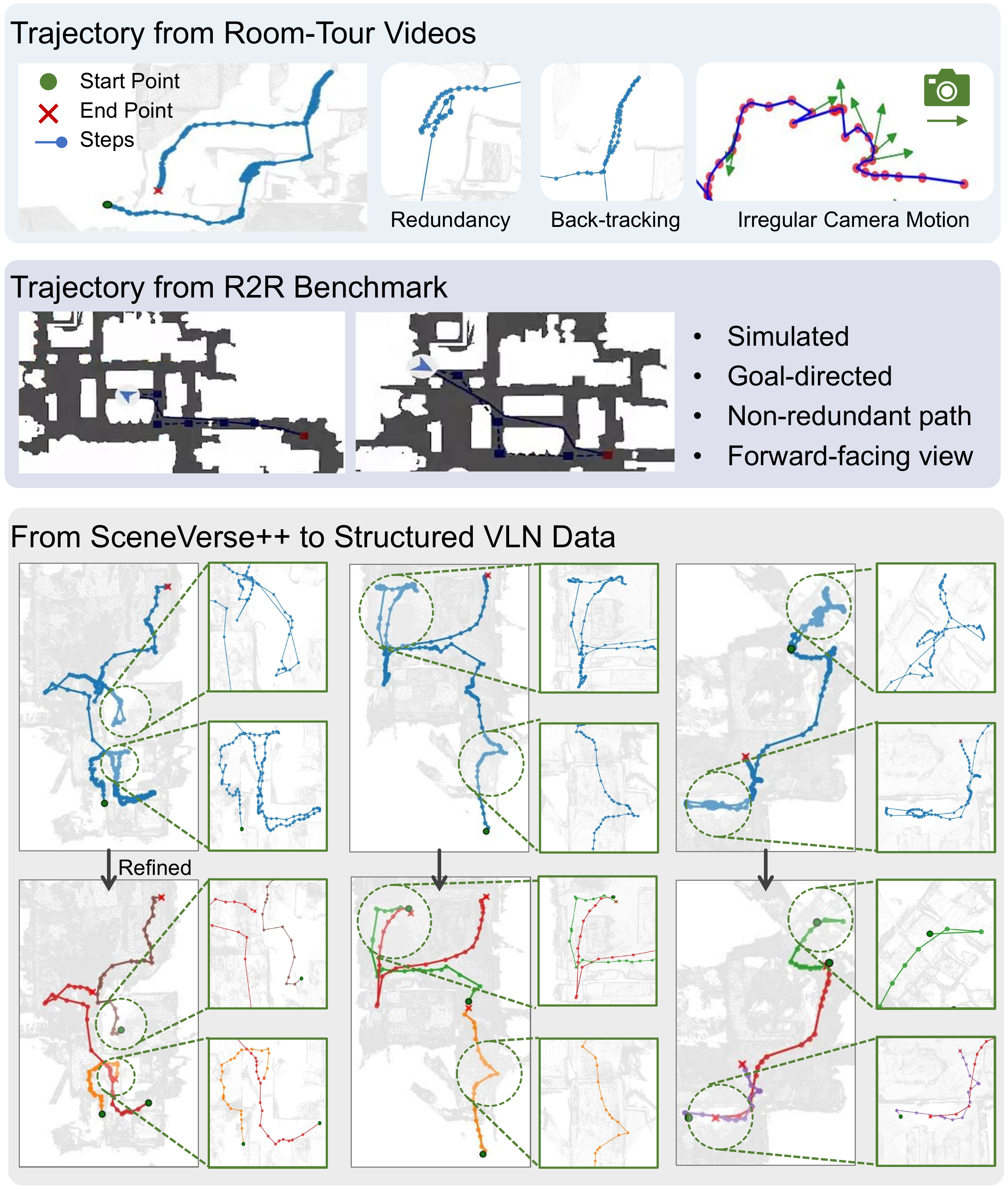}
    \caption{\textbf{Trajectory comparison.} Top: Room-tour videos show irregular and redundant camera motions. Middle: R2R trajectories are smooth and goal-directed. Bottom: raw videos are converted into VLN-compatible data. Different colors indicate sub-paths.}
    \vspace{-3mm}
    \label{fig:vln_data}
\end{figure}

\paragraph{Data Generation}
R2R establishes a controlled and standardized setting for instruction-following navigation, but its simulated trajectories differ from how humans naturally explore real environments.
Specifically, VLN trajectories are goal-directed shortest paths with all forward-facing movements, whereas room-tour videos capture free-form exploration in the environment, often exhibiting \textit{irregular camera motion}, \textit{redundancy}, and \textit{backtracking}, shown in \cref{fig:vln_data}.
These discrepancies introduce challenges in deriving navigation-consistent trajectories from room-tour videos for VLN model learning. To bridge this gap, we analyze human motion patterns in real videos and design a three-stage pipeline to convert room-tour camera trajectories to navigation trajectories as in \cref{fig:vln_system}.

\textit{Path Pre-processing.} 
The goal of this stage is to extract clean and coherent trajectories from room-tour videos using the \ac{sfm} reconstructions described in \cref{sec:curation}.
We first cluster camera positions within a 0.5m radius to merge nearby viewpoints, keeping one representative node per cluster to maintain trajectory continuity and remove redundant local rotations.
Next, to eliminate backtracking, we split long trajectories into sub-paths. Specifically, we detect cluster centers along each trajectory and use them as potential break points—only when the two adjacent segments separated by a center exceed 15 steps do we perform a split.
Finally, we filter out steps that involve rotations greater than 90° or translations larger than 70cm.
The resulting trajectories are shown at the bottom of \cref{fig:vln_data}.

\textit{Action Encoding.} 
Action encoding converts each structured trajectory into a sequence of discrete actions for VLN model training. 
We extract each node’s 3D camera pose $(\mathbf{R}_i, \mathbf{t}_i)$ from the SfM reconstruction, project it onto the ground plane, and represent it as
$p_i = [x_i,\, y_i,\, \theta_i]$,
where $(x_i, y_i)$ denotes the position and $\theta_i$ is the yaw angle derived from $\mathbf{R}_i$.
As room-tour videos often contain irregular camera motions, we remove non-navigational ``looking around motions'' by removing actions whose viewing direction deviates from walking direction.
Finally, the movement action is defined by the Euclidean distance between $p_i$ and $p_{i+1}$, and rotation action by $\Delta \theta_i = \theta_{i+1} - \theta_i$. 
Both are discretized following the R2R convention for compatibility with existing VLN benchmarks.

\textit{Instruction Generation.} 
We leverage \ac{vlm} to generate natural language navigation instructions aligned with both motion and visual context, by providing both the corresponding images and encoded actions. 
The \ac{vlm} first reasons about local motion changes using \ac{cot} and then composes coherent instructions to describe the entire trajectory. To enhance linguistic diversity and improve generalization, we generate three stylistically varied instructions for each trajectory.

\paragraph{Statistics}
The \ac{vln} data derived from \dataset contains 9,631 trajectories, each averaging 12.8 meters in length and 15 steps. For each trajectory, we provide three instructions in formal, conversational, and narrative styles, averaging 42, 47, and 57 words, respectively. After discrete action encoding, forward and rotational movements account for 52\% and 48\%, reflecting a balanced motion distribution.
R2R comprises 7,189 trajectories and 21,567 instructions averaging 29 words, collected from 29 simulated indoor scenes. 
Our dataset extends these benchmarks by incorporating richer linguistic diversity and natural, real-world motion patterns captured from internet videos.

\paragraph{Performance}
We validate the effectiveness of our constructed VLN dataset using LLaVA-Video~\cite{zhang2024video} as the base model. All experiments are evaluated on the validation set of R2R~\cite{anderson2018vln}. To ensure a fair comparison, we use the same number of training epochs across all experiments.
Evaluation follows standard VLN metrics, including \textit{Distance to Goal} (Dist.), \textit{Success Rate} (SR), \textit{Oracle Success} (OS), \textit{Success-weighted Path Length} (SPL) and \textit{Path Length} (PL).

\begin{itemize}
    \item \textbf{Domain Transfer and Training Strategies.} 
    We investigate how incorporating real-world video data affects \ac{vln} performance on R2R. As shown in \cref{tab:vln_llava_netsun}, training on \dataset yields a modest SR improvement (0.107 \vs 0.088) under zero-shot evaluation compared with training on R2R alone. The substantially longer path length (14.1 \vs 5.2) reflects that the richer and more complex trajectories in room-tour videos offer more diverse and challenging experiences to learn navigation behaviors, compared with the shortest paths in R2R.
    Further fine-tuning on R2R significantly boosts SR to 0.228, demonstrating that large-scale video pretraining provides valuable visual and linguistic priors for navigation tasks.
    In contrast, directly mixing \dataset with R2R during training yields weaker results, suggesting that the visual gap between real videos and simulator-rendered scenes makes naive mix-training less effective.
    
    \item \textbf{Data Quality.}  To further investigate the impact of data quality on VLN performance, we conduct ablation experiments on two core components in data generation: trajectory refinement (TR) and instruction enrichment (IE).
    As shown in \cref{tab:vln_llava_netsun}, removing either component results in a clear performance drop.
    Even after fine-tuning on R2R, models pretrained on these ablated datasets fail to fully recover, \eg, SR decreases from 0.228 to 0.177 when TR is removed.
    These results demonstrate that raw internet videos alone are insufficient for effective VLN training; task-specific data processing is essential.
    We additionally include comparisons with YouTube-based VLN data from NaVILA~\cite{cheng2024navila} in \supp, where \dataset enables stronger model performance due to higher quality.
\end{itemize}

\begin{table}[t!]
    \centering
    \caption{\textbf{VLN evaluation under different training settings}. TR denotes trajectory refinement, and IE for instruction enrichment.}
    \label{tab:vln_llava_netsun}
    \resizebox{\linewidth}{!}{
        \begin{tabular}{cc|ccccc}
            \toprule
            \textbf{Pretrain} & \textbf{Finetune} & 
            \textbf{SR↑} & \textbf{OS↑} & \textbf{SPL↑} & \textbf{Dist↓} & \textbf{PL} \\
            \midrule
            - & R2R & 0.088 & 0.133 & 0.076 & 8.031 & 5.222 \\
            R2R + \dataset   & - & 0.188 & 0.262 & 0.150 & 8.117 & 10.496\\
            \dataset  & –            & 0.107 & 0.194 & 0.074 & 9.418 & 14.097 \\
            \dataset  & R2R          & \textbf{0.228} & \textbf{0.315} & \textbf{0.191} & \textbf{7.65} & 11.642 \\
            \midrule
            \dataset (w/o IE) & – & 0.022 & 0.043 & 0.016 & 8.978 & 2.333 \\
            \dataset (w/o IE) & R2R & 0.074 & 0.111 & 0.062 & 8.175 & 5.009 \\
            \dataset (w/o TR) & – & 0.036 & 0.045 & 0.032 & 8.662 &2.521 \\
            \dataset (w/o TR) & R2R & 0.177 & 0.298 & 0.130 & 8.23 & 11.949 \\
            
            \bottomrule
        \end{tabular}
    }
\end{table}

\section{Discussion and Conclusion}
In this paper, we investigate pathways to advance comprehensive 3D scene understanding across multiple tasks by leveraging unlabeled internet videos. We develop automated data engines to generate training data and demonstrate that high-quality data can benefit downstream tasks. We further offer the following discussions on data generation, benchmarks, and model scaling. Limitations and future work are discussed in \supp.

\paragraph{Scaling capability of models.} In our experiments, we observe clear differences in how models scale. Models that depend on task-specific, pre-computed segments are more sensitive to data distribution shifts and hyperparameter changes, resulting in limited scalability and weaker generalization in 3D instance segmentation (\cref{sec:segmentation}). In contrast, models that operate directly on raw and widely available modalities, \eg, 3D voxels or RGB-based \acp{mllm}, exhibit more robust scaling behavior. This contrast is less evident in two-dimensional settings due to the uniformity of image inputs, but becomes increasingly pronounced when scaling 3D understanding.

\paragraph{Fair evaluation of capability and benchmarks.} Existing benchmarks may not fully reflect a model's true capability, \eg, VSI-Bench exhibits strong \ac{qa} distribution bias~\cite{brown2025sims} and \acp{vlm} overfit to data-specific cues for in-domain evaluation (\cref{sec:3dvqa}). To ensure fair assessment, future evaluation should emphasize zero-shot testing on existing benchmarks, avoiding data contamination and minimizing data distribution gap, or more benchmarks that accurately measure 3D scene understanding and generalization in the wild.

\paragraph{Understanding data and task-specific biases.} Effective data scaling requires not only high-quality data, but also a careful examination of data distribution and task-specific or benchmark-specific characteristics. Performance is strongly affected by factors that remain hidden without deeper analysis, \eg, the discrepancy between natural camera motion in real-world videos and goal-directed navigation trajectories (\cref{sec:navigation}). Identifying such mismatches is essential to avoid biases and to ensure that scaled data provides meaningful improvements for the intended task.

\paragraph{Advancing automated data generation.}
Building an automated data generation pipeline reveals significant challenges in using existing models to produce high-quality data for 3D scene understanding from in-the-wild videos. Modules such as \ac{sfm}, instance segmentation, and language grounding are typically trained on task-specific or small-scale benchmarks, limiting their generalization capabilities and introducing sequential errors when combined together for in-the-wild spatial understanding. As a result, substantial effort is required for careful model selection and non-trivial coordination across modules. We advocate that future development of these sub-modules should align with the broader goal of enabling robust in-the-wild 3D understanding, with evaluation based not only on task-specific performance but also on their contribution to reliable automated data generation pipelines.


{
    \small
    \bibliographystyle{ieeenat_fullname}
    \bibliography{main}
}

\clearpage
\appendix
\setcounter{page}{1}
\setcounter{table}{0}
\setcounter{figure}{0}
\setcounter{footnote}{0}
\renewcommand{\thefigure}{S.\arabic{figure}}
\renewcommand{\thetable}{S.\arabic{table}}
\renewcommand{\theequation}{S.\arabic{equation}}
\maketitlesupplementary

\section{Data Curation}

\begin{figure*}[!t]
    \centering
    \includegraphics[width=1.0\linewidth]{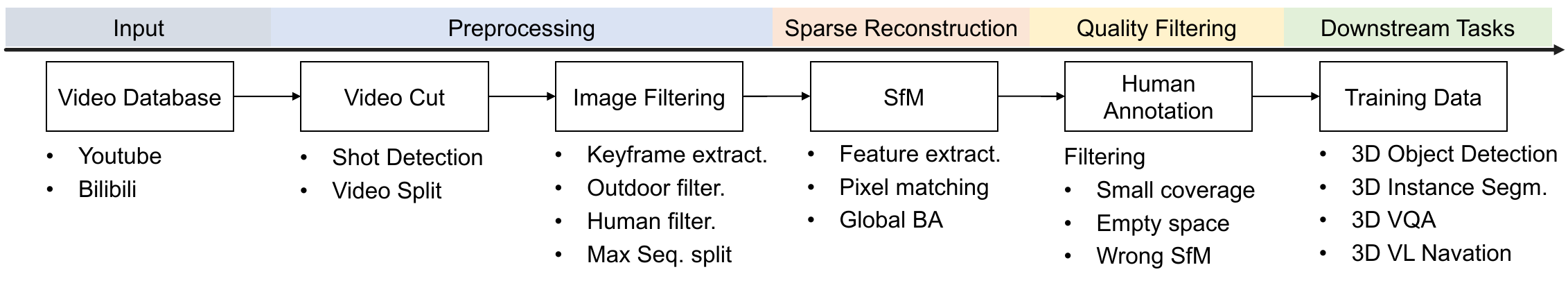}
    \caption{\textbf{Overview of data curation pipeline}.}
    \label{fig:data curation}
\end{figure*}

As described in \cref{sec:curation}, we provide detailed information on how sparse reconstruction data are generated from Internet videos. The raw Internet data are collected from housing‐tour videos on YouTube\footnote{\url{http://youtube.com/}} and Bilibili\footnote{\url{http://bilibili.com/}}, which contain a total of 8,217 videos, from which we obtain 6,687 reconstructed scene instances. The overall data processing pipeline consists of two main stages: preprocessing and reconstruction, as shown in \cref{fig:data curation}.

\paragraph{Preprocessing stage} Internet videos often contain many shots rather than a continuous shot, which can significantly degrade the reconstruction quality if treated as a single sequence. To address this, we use TransNetV2~\cite{soucek2020transnetv2} to detect the shot boundary and split each video into multiple shots, each treated as an individual scene. Since each clip still includes a large number of redundant or noisy frames, we use parallax-based keyframe selection to retain representative frames, and employ detection models to filter out outdoor frames and frames that contain humans. To ensure both reconstruction efficiency and quality, long sequences are further subdivided based on the number of keyframes, with a maximum clip length of 300 frames and an overlap of 50 frames between adjacent clips.

\paragraph{Reconstruction stage} To efficiently establish image correspondences, we use a loop and sequence pairing strategy. In the loop pairing strategy, we extract image features and compute feature distances to other images within a 100‐frame range. The top 50 image pairs with feature distances greater than 0.4 are retained as valid loop pairs. In the sequence pairing strategy, the preceding and following 20 frames are used as sequential pairs. We then extract feature points~\cite{duisterhof2025mast3r} for each image pair and perform feature matching across the selected pairs to generate point correspondences. Finally, we use COLMAP to estimate the camera parameters and complete the sparse reconstruction.

\begin{figure}[t]
    \centering
    \begin{subfigure}{\linewidth}
        \centering
        \includegraphics[width=1.0\linewidth]{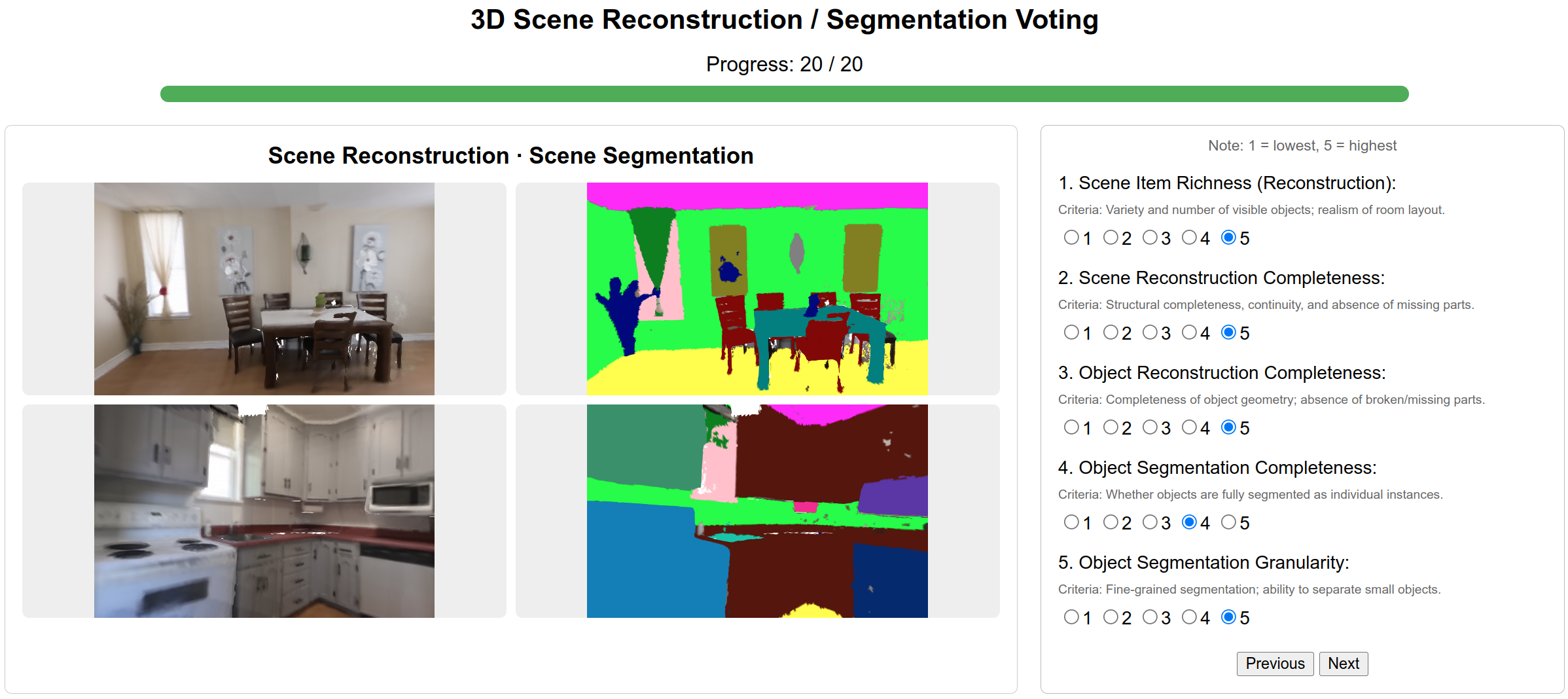}
        \caption{An example from \dataset.}
    \end{subfigure}
    \begin{subfigure}{\linewidth}
        \centering
        \includegraphics[width=1.0\linewidth]{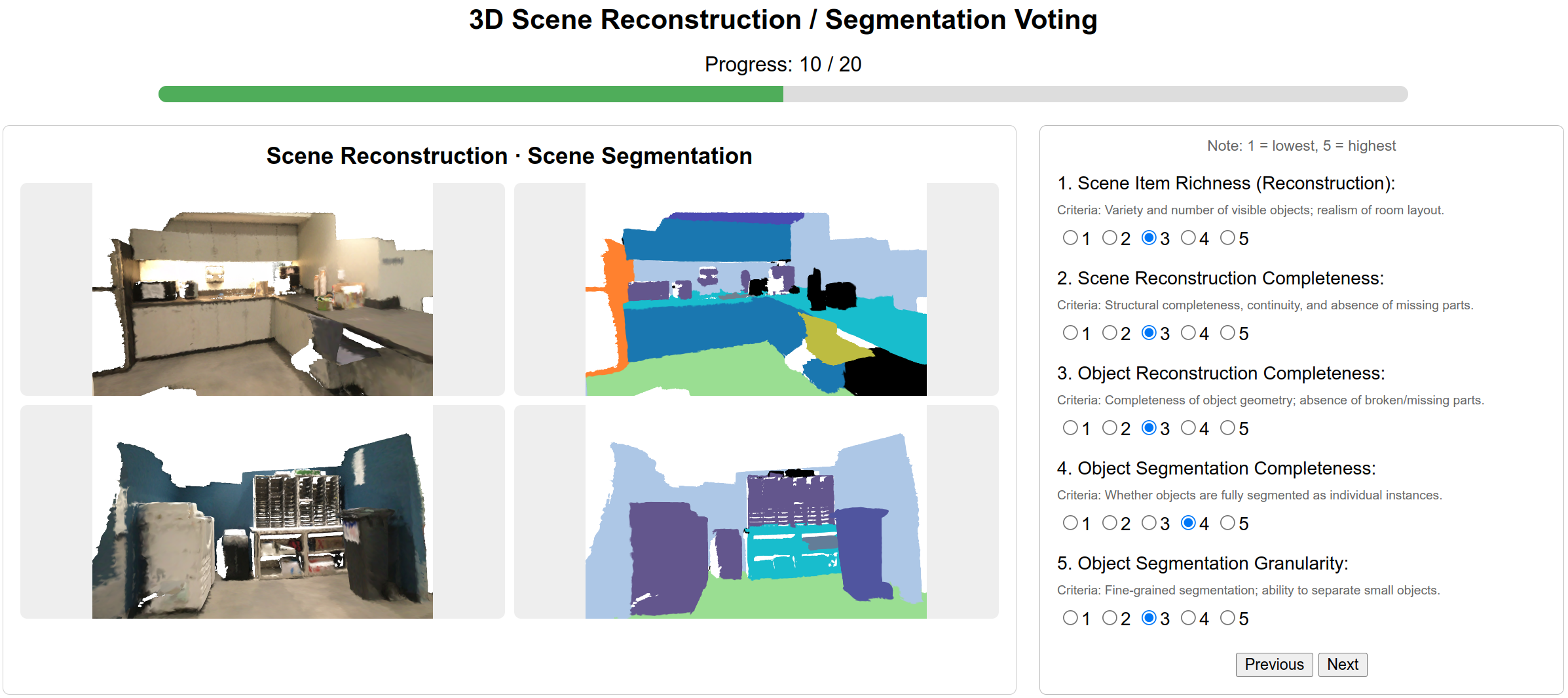}
        \caption{An example from ScanNet.}
    \end{subfigure}
    \caption{\textbf{Example of quality check.} The data samples from \dataset and ScanNet are mixed and anonymous.}
    \label{fig:vote_system}
\end{figure}
\section{Data Quality Check}
To assess the quality of data produced by our automated data engine, we perform a human evaluation on the reconstruction and instance segmentation. More specifically, we sample 10 scenes from \dataset and ScanNet, respectively, visualize their reconstruction and segmentation results side-by-side, and ask human subjects to rate each scene on a scale of 1 to 5, along the following 5 axes:
\begin{itemize}
    \item \textbf{Scene Item Richness}: diversity of abundance of visible items, and how well they reflect realistic indoor layouts.
    \item \textbf{Scene Reconstruction Completeness}: structural completeness of the reconstructed scene, including coherence and absence of major holes or missing regions.
    \item \textbf{Object Reconstruction Completeness}: integrity of individual object shapes, with no breaks, missing faces, or lost components.
    \item \textbf{Object Segmentation Completeness}: whether each object is segmented as a single, coherent instance without obvious omissions or incorrect splits.
    \item \textbf{Object Segmentation Granularity}: the fineness of segmentation, segmenting small objects accurately and avoiding unintended merging.
\end{itemize}
The results are in \cref{tab:supp_quality}. From the table, \dataset achieves quality comparable to or exceeding ScanNet across the above evaluation criteria, especially in the richness and completeness of reconstruction, which shows that our dataset captures diverse and real-world distributions. It also indicates that modern image-based reconstruction and segmentation methods, if properly adapted, have advanced to a point where they can surpass the sensor quality and reconstruction pipeline used in ScanNet capture in 2017. This highlights their potential for further scaling. The grading interface is shown in \cref{fig:vote_system}.

\begin{table}[t!]
\caption{\textbf{The quality check of \dataset and ScanNet.}}
\centering
\small
\begin{tabular}{lcc}
\toprule
\textbf{Criterion} & \textbf{\dataset} & \textbf{ScanNet} \\
\midrule
Scene Item Richness & 4.43 & 3.68 \\
Scene Reconstruction Completeness & 4.25 & 3.09 \\
Object Reconstruction Completeness & 4.16 & 3.23 \\
Object Segmentation Completeness & 3.93 & 3.26 \\
Object Segmentation Granularity & 3.89 & 3.24 \\
\midrule
Average & 4.13 & 3.30\\
\bottomrule
\end{tabular}
\label{tab:supp_quality}
\end{table}

\section{Experiment Details}

\subsection{3D Object Detection and Segmentation}

\paragraph{3D Object Detection} 
To better handle the large scenes in \dataset, we adopt an additional spatial cropping augmentation during SpatialLM training. For each sample, one object is randomly selected, and the point cloud within a 3-meter radius of the object is extracted and used as the model input.
SpatialLM is trained on \dataset using 8 NVIDIA A100 GPUs for 1000 epochs with a batch size of 1, requiring approximately 2 days. The model is then fine-tuned on ScanNet for another 1000 epochs with a batch size of 4, which takes about 12 hours. For supervision, we utilize 15 semantic categories selected from the ScanNet 20 labels.

\paragraph{3D Instance Segmentation}

\begin{figure*}
    \centering
    \includegraphics[width=1.0\linewidth]{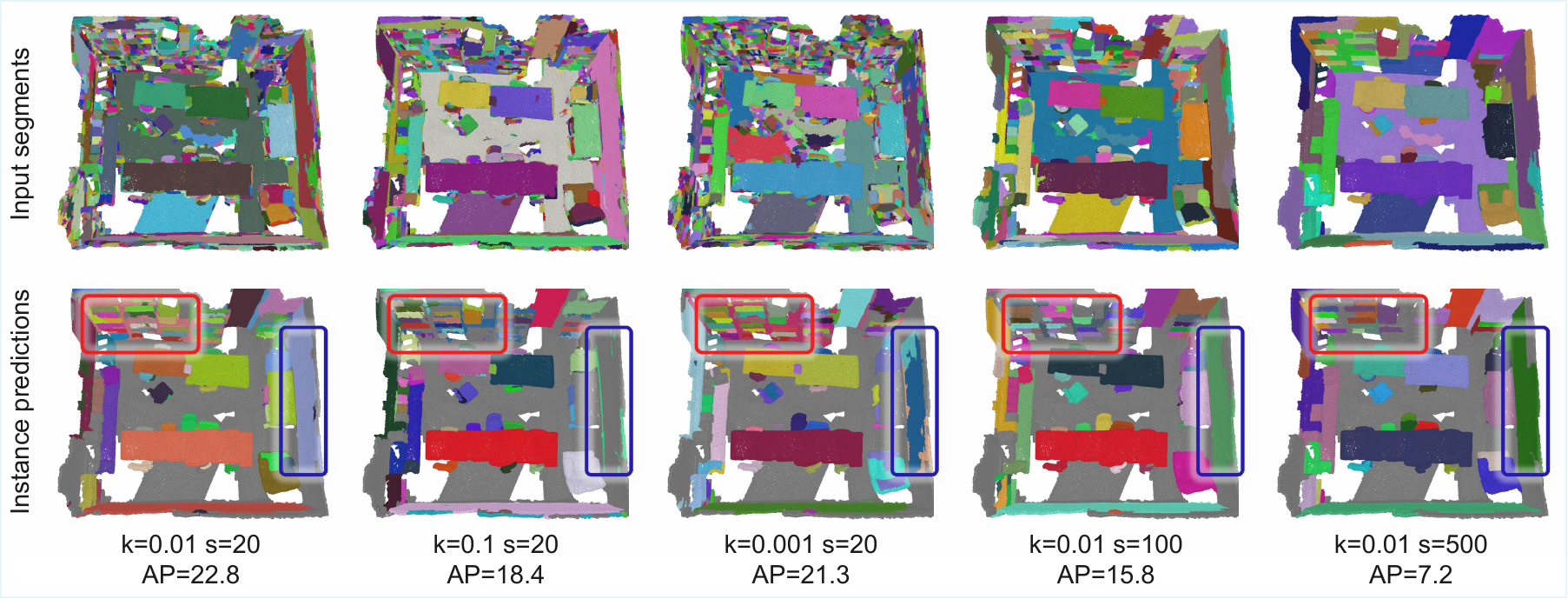}
    \caption{\textbf{Qualitative example of sensitivity tests on different segment distributions.} We evaluate the model trained with k=0.01 and s=20 on other segment settings. The first row shows the input segments, and the second row shows the 3D instance prediction. As the segments become smaller, over-segmentation gradually appears (highlighted by the blue boxes). Conversely, as the segments become larger, under-segmentation becomes increasingly evident (see the red boxes). The AP is reported as the average over the whole ScanNet.}
    \label{fig:diff_segments}
\end{figure*}

In 3D instance segmentation experiments, we observe that the model trained on \dataset does not transfer well to ScanNet. One important reason is that Mask3D relies on the segment-level masks produced by a graph-based segmentation method, and different hyperparameters lead to noticeably different segment results. Two decisive hyperparameters, segmentation threshold (kThresh) and minimum segment size (segMinVerts), directly control the connectivity and granularity of segments. To illustrate this sensitivity, we provide further experiments by evaluating a model trained on ScanNet (with kThresh=$10^{-2}$ and segMinVerts=20), on segments generated from different hyperparameter settings. As shown in \cref{tab:segments_example} and \cref{fig:diff_segments}, coarse segments fail to correctly isolate individual instances, while overly fine segments result in fragmented predictions and miss detections. This issue is more obvious during the training stage, where the mismatched segment distribution causes poor model transfer. These observations highlight a broader challenge in scaling 3D scene understanding: models sensitive to task-specific modalities and data distribution shifts exhibit limited scalability, whereas models operating directly on raw and widely available modalities may scale more robustly.

\begin{table}[!t]
\centering
\small
\caption{\textbf{Evaluation sensitivity on different segment settings on 3D instance segmentation.} The model is trained with kThresh=$10^{-2}$ and segMinVerts=20, and performance degrades if the distribution of the testing segments diverges from training. }
\label{tab:segments_example}
\begin{tabular}{cc|ccc}
\toprule
\textbf{kThresh} & \textbf{segMinVerts} & \textbf{AP$_{25}$} & \textbf{AP$_{50}$} & \textbf{AP}\\
\midrule
$10^{-2}$ & 20   & 36.1 & 31.8 & 22.8 \\ 
\midrule
$10^{-1}$ & 20   & 34.6 & 28.1 & 18.4 \\ 
$10^{-3}$ & 20   & 35.9 & 30.4 & 21.3 \\ 
$10^{-2}$ & 100  & 30.8 & 24.6 & 15.8 \\ 
$10^{-2}$ & 500  & 17.9 & 12.6 &  7.2 \\ 
$10^{-1}$ & 1000 & 11.2 &  7.7 &  4.2 \\ 
$10^{-3}$ & 1000 & 10.9 &  7.5 &  4.1 \\ 
\bottomrule
\end{tabular}
\end{table}

\subsection{3D Spatial VQA}
\paragraph{Data Generation}
From the 3D reconstruction and instance segmentation results, we first construct the overall per-scene information, \ie, the room size. Next, we automatically construct 3D scene graphs from point clouds. We first instantiate the graph nodes with the instance annotation from the point cloud and parameterize each node with the object centroid and size of the axis-aligned bounding box. Next, we traverse all the nodes to determine their spatial relationships, following \citet{jia2024sceneverse}. We then save the counts for different object categories and generate the \acp{qa} accordingly.
\begin{itemize}
    \item Object Count (\ac{na}): Count the number of instances of a specified object category that has more than 1 instance within a room.
    \item Relative Distance (\ac{mca}): Identify which of four candidate objects is closest in 3D space to a target object, which can be uniquely identified by its category.
    \item Relative Direction (\ac{mca}): Given a situation describing the observer's position and orientation, determine the relative direction of a query object, which can be uniquely identified by its category.
    \item Object Size (\ac{na}): Estimate the length of the longest dimension of an object instance in centimeters.
    \item Absolute Distance (\ac{na}): Estimate the Euclidean distance between the closest points of two specified objects in meters. The two objects are randomly selected from the categories that have only one instance.
    \item Room Size (\ac{na}): Estimate the area of the room in square meters (numerical answer).
    \item Route Planning (\ac{mca}): We generate \ac{qa} pairs by employing the navigation trajectories within 3D environments in the \ac{vln} task. The actions are masked to create multiple-choice, and the navigation summary is transferred to guidance via \ac{vlm}. Detailed prompts are in \cref{tab:rp_vqa_prompt_1,tab:rp_vqa_prompt_2}.
\end{itemize}

\paragraph{Dataset Statistics}
Applying the generation pipeline to \dataset yields 632K spatial VQA data following the VSI-Bench format. It comprises 391K samples for  \ac{mca} and 241K samples for \ac{na} with 7 different question types. The number of each type of question is listed in \cref{tab:supp_vqa_stats}. In our experiment, we sampled a subset of 202K for training.
\begin{table}[ht]
\caption{\textbf{3D Spatial VQA Data Distribution.}}
\centering
\small
\begin{tabular}{lr}
\toprule
\textbf{Task Type} & \textbf{Count} \\
\midrule
Object Relative Direction & 155,199 \\
Object Absolute Distance & 137,397 \\
Object Relative Distance & 226,639 \\
Object Size Estimate & 44,050 \\
Object Count & 53,200 \\
Route Plan & 9,588 \\
Room Size & 6,684 \\
\midrule
\textbf{Total} & \textbf{632,757} \\
\bottomrule
\end{tabular}
\label{tab:supp_vqa_stats}
\end{table}

\paragraph{Training Configuration}
All experiments for 3D VQA fine-tuning were conducted using LoRA-based adaptation on an LLM backbone, with training performed on 4 × NVIDIA A100 GPUs. More Training Configuration and Reproducibility Details are provided in \cref{tab:supp_vsq_training_details}.
\begin{table}[ht]
\centering
\caption{\textbf{Training Details for 3D Spatial VQA.}}
\label{tab:supp_vsq_training_details}
\resizebox{\linewidth}{!}{
\begin{tabular}{l l}
\toprule
\textbf{Category} & \textbf{Setting} \\
\midrule
Hardware & 4 $\times$ NVIDIA A100 GPUs \\
Precision & BF16 \\
\midrule
LoRA Rank & 128 \\
LoRA Scaling Factor & 256 \\
\midrule
Per-device Batch Size & 1 \\
Gradient Accumulation Steps & 32 \\
Effective Batch Size & $4 \times 32 = 128$ \\
\midrule
Optimizer & AdamW \\
Learning Rate & $2\times10^{-5}$ \\
Weight Decay & 0 \\
Warmup ratio & 0.03 \\
LR Schedule & cosine \\
\midrule
Epochs & 5 \\
Actual Training & Stop after 1 epoch \\
\midrule
Random Seed & 42 \\
\bottomrule
\end{tabular}
}
\end{table}

\subsection{3D \texorpdfstring{\acf{vln}}{}}
\paragraph{Depth Scale Calibration}
We design a three-stage pipeline to convert room-tour camera trajectories to VLN trajectories in \cref{sec:navigation}. 
In action encoding stage, we apply a scale calibration procedure during the action-encoding stage to ensure that movement distances computed from trajectories reflect real-world scale. This is necessary because the SfM reconstruction provides depth on an arbitrary scale, whereas VLN models require physically meaningful forward-motion distances.
To estimate the correct scale, we identify video frames containing large and visually stable furniture (e.g., sofas, cabinets, refrigerators), whose depths are easier to estimate reliably. For each selected region, we obtain a robust monocular depth estimate using Depth-Pro~\cite{bochkovskiy2025depth}. In parallel, we extract the corresponding absolute (but unscaled) depth from the SfM reconstruction. By comparing these two depth values, we compute a depth-scale factor for each furniture instance. The scale factors are averaged across all selected samples to produce a stable calibration value, which is then applied to the entire reconstructed scene.
Accurate depth calibration ensures that forward-motion distances derived from trajectories correspond to realistic navigation steps, improving the reliability of action encoding for VLN training. The prompt used for instruction generation is provided in \cref{tab:instruction_prompt}.

\paragraph{Training Configuration}
We train LLaVA-Video on 8 NVIDIA A100 GPUs. Zero-shot and mixed-training experiments are run for 1 epoch, while the pretrain–finetune setting uses 2 epochs of pretraining on \dataset and 1 epoch of fine-tuning on R2R. To ensure balanced exposure to actions across datasets, we apply label rebalancing: we count the occurrences of each action category across all episodes and select a reference frequency based on the median or maximum count. Actions below the reference are oversampled, and actions above the reference are subsampled. Finally, we use the total number of R2R training samples as the baseline and adjust other datasets accordingly to maintain comparable sample counts. Each epoch of training takes approximately 14 hours with a batch size of 2.

\paragraph{Comparison with Internet-Scale VLN Data}
To validate the effectiveness of our \dataset, we compare it with the YouTube-derived VLN data from NaVILA~\cite{cheng2024navila}, which contains roughly 20k trajectories, using Qwen-VL-7B~\cite{bai2025qwen2} as the base model. We evaluate two settings, zero-shot and mixed-training with R2R, and report results in \cref{supp:tab:vln_llava_netsun}.
In the zero-shot setting, \dataset and NaVILA show similar performance (SR = 0.09). \dataset exhibits a larger path length (PL = 11.274), reflecting the inherently longer trajectories present in our data generation pipeline.
In the mixed-training setting, \dataset yields clear improvements over NaVILA on key navigation metrics: higher Success Rate (0.32 vs. 0.29), higher SPL (0.258 vs. 0.213), and lower Distance-to-Goal (7.447 vs. 7.960). This indicates that \dataset provides more effective supervision for learning grounded navigation when combined with R2R.

Notice that NaVILA contains roughly 2.5× more data than \dataset, which may bias certain metrics in its favor. Despite this scale advantage, \dataset still achieves superior SR and SPL, suggesting that well-structured, navigation-aligned trajectories are more beneficial than raw data volume alone. These findings support the value of our data-generation pipeline while also underscoring the need to further explore domain differences and dataset scaling in future work.

\begin{table}[t!]
    \centering
    \caption{\textbf{Comparison between NaVILA and \sdataset.} Experiments use Qwen2.5-VL-7B under zero-shot and mixed-training.}
    \label{supp:tab:vln_llava_netsun}
    \resizebox{\linewidth}{!}{
    \begin{tabular}{l c | c c c c c}
        \toprule
        \textbf{Data Source} & \textbf{Setting} &
        \textbf{SR↑} & \textbf{OS↑} & \textbf{SPL↑} & \textbf{Dist↓} & \textbf{PL} \\
        \midrule
        NaVILA  & Zero-shot & \textbf{0.09} & 0.132 &\textbf{0.08}  & \textbf{9.406} & 8.505 \\
        \sdataset  & Zero-shot & \textbf{0.09} & \textbf{0.145} & 0.063 & 9.439 & 11.274 \\
        \midrule
        R2R + NaVILA  & Mix & 0.29 & \textbf{0.424} & 0.213 & 7.960 & 16.013 \\
        R2R + \sdataset  & Mix & \textbf{0.32} & 0.402 & \textbf{0.258} & \textbf{7.447} & 12.918 \\
        \bottomrule
    \end{tabular}
    }
\end{table}
\begin{figure}[ht!]
    \centering
    \includegraphics[width=\linewidth]{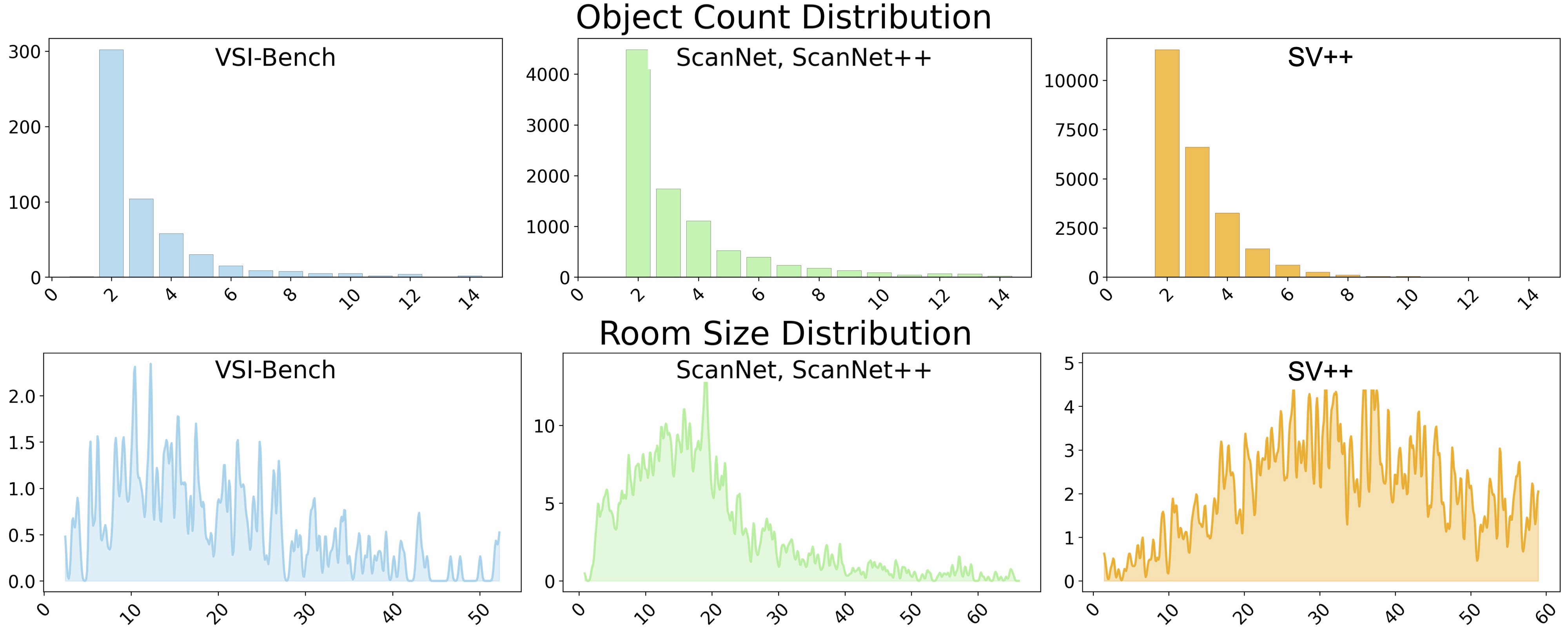}
    \caption{\textbf{3D spatial VQA answer distribution.}}
    \label{supp:fig:distribution_compare}
\end{figure}

\section{More Discussion}
\paragraph{Why ``Object Count'' and ``Room Size'' performance drop in 3D spatial VQA?} We believe the data distribution bias is the major factor here. Several pieces of evidence: 1) \dataset and ScanNet/ScanNet++ GT perform similarly on zero-shot experiment in Tab. 3; 2) From \cref{supp:fig:distribution_compare}, Object Count test distribution in VSI-Bench is highly biased at ``2'', where in-domain data (ScanNet / ScanNet++) has a much smaller divergence to this peak, showing potential benchmark overfitting:
\begin{align*}
D^{obj\_cnt}_{KL}(\text{VSI-Bench} \parallel \text{\dataset})=1.04 \\
D^{obj\_cnt}_{KL}(\text{VSI-Bench} \parallel \text{SN,SN++})=0.145.
\end{align*}
Room size shows a larger domain gap:
\begin{align*}
D^{room\_size}_{KL}(\text{VSI-Bench} \parallel \text{\dataset})=6.08 \\
D^{room\_size}_{KL}(\text{VSI-Bench} \parallel \text{SN,SN++})=2.95, 
\end{align*}
where \dataset signatures multi-room scenes. 

\begin{figure}[h]
     \centering
     \begin{subfigure}[t]{0.48\linewidth}
         \centering
         \includegraphics[width=\textwidth]{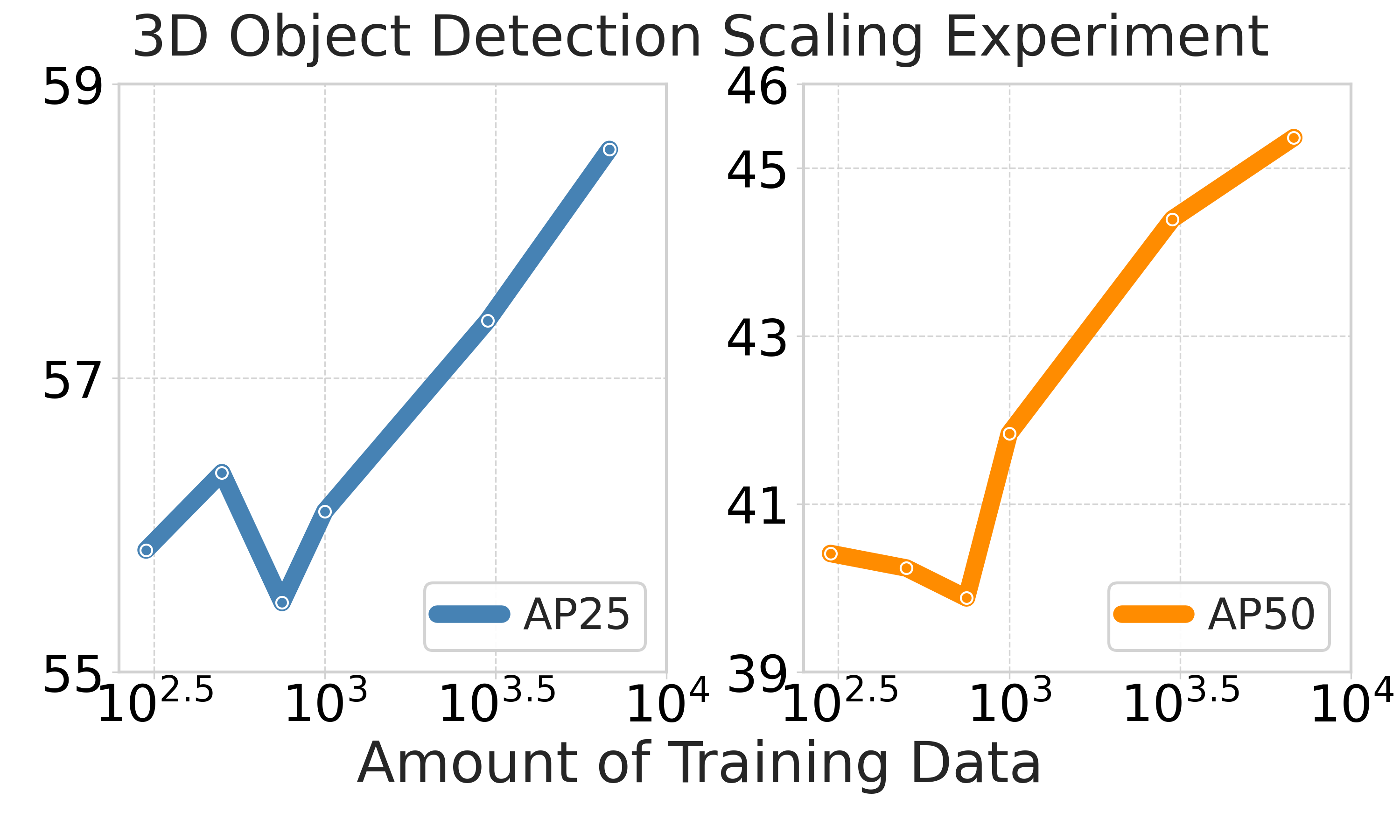}
     \end{subfigure}
     \begin{subfigure}[t]{0.48\linewidth}
         \centering
         \includegraphics[width=\textwidth]{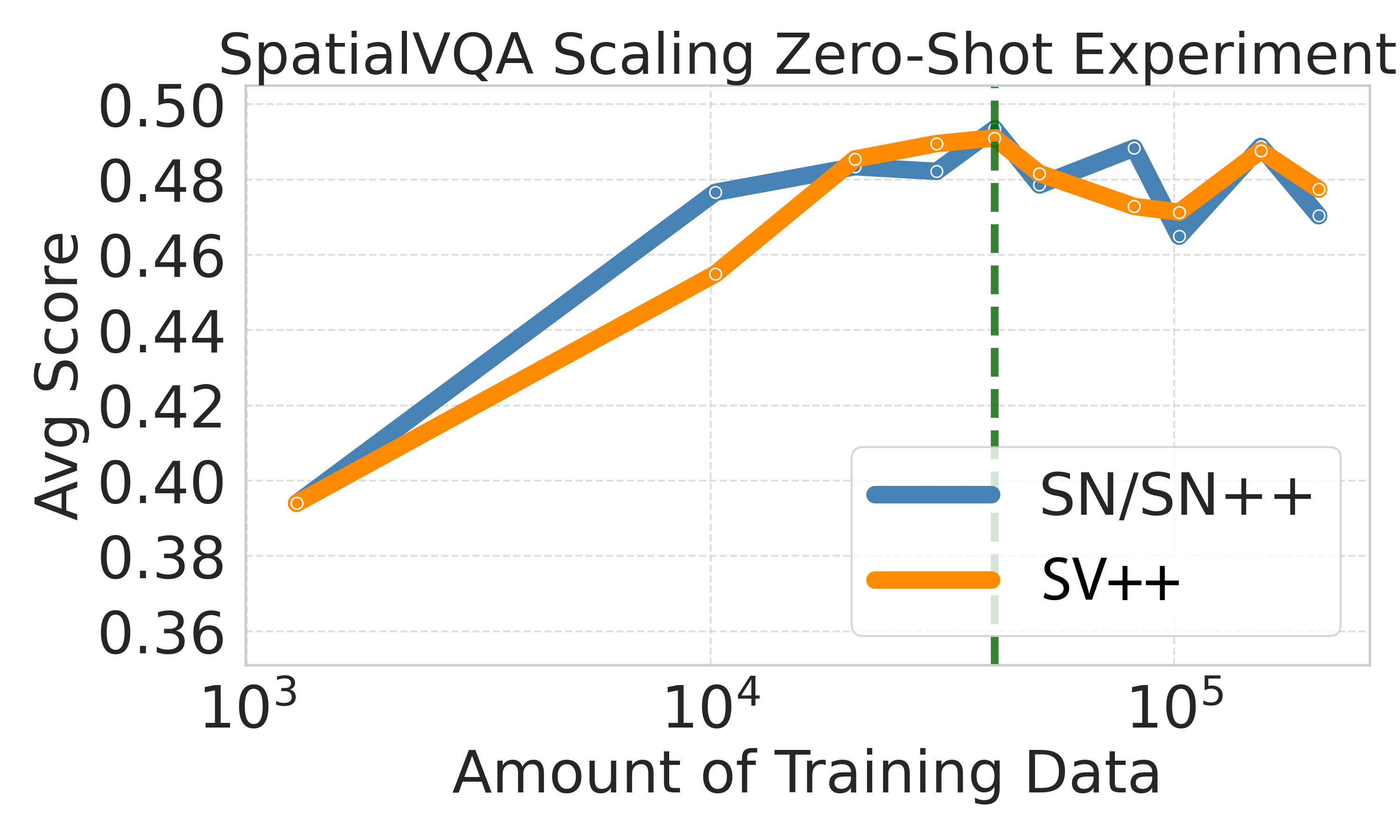}
     \end{subfigure}
     \caption{\textbf{Data scaling effects.}}
     \label{supp:fig:data_scaling}
\end{figure}

\paragraph{Data scaling analysis} We provide scaling results for 3D detection and 3D VQA in \cref{supp:fig:data_scaling}, where $\text{data} \sim \mathcal{O}(N_{\text{scenes}})$. Performance follows a \textit{log-linear trend} in both cases, but VQA reaches saturation later. More effective scaling requires co-design involving model architecture, fair benchmarks, and data quality.

\paragraph{Per-scene computation overhead} The average end-to-end per-scene running time is $\sim$0.59h, consisting of 0.27 GPU-hours (RTX 3090-level) and 0.32h CPU-hour (Xeon 14 vCPUs). Stage-wise, preprocessing and SfM take 69.8\%, depth and 2D segmentation model inference takes 23.2\%, dense 3D reconstruction 3\% and 3D segmentation 4\%. This overhead is manageable for large-scale data generation and could be further optimized.

\section{Limitations and Future Work} 
Limited by computational resources, our experiments are bound to the minimal setting to examine the contribution of different data sources. In practice, 3D understanding capability also depends on the base model capacity, optimization strategy, and data mixture, \eg, existing \ac{vln} systems often benefit from larger training corpora. Additionally, internet videos may contain privacy-sensitive content from public areas. Scaling such data requires careful adherence to ethical guidelines, regulatory frameworks, and responsible development. Future work includes iterative refinement of the generated data, integration with more advanced models to further enhance capability, and extending to dynamic videos that capture the 4D scene evolution.

\begin{table*}[t]
\centering
\caption{\textbf{Prompts for Navigation instruction generation in \dataset.}}
\small
\label{tab:instruction_prompt}
\resizebox{\linewidth}{!}{
\begin{tabular}{p{0.95\linewidth}}
\toprule
You are an embodied AI agent making task summaries for a navigation task. Your goal is to generate faithful, human-readable navigation instructions.

\textbf{--- Input ---}\\[-6pt]
\begin{itemize}
    \item A sequence of first-person images and a stepwise action sequence.
    \item Each image corresponds to the visual observation immediately before the action in that frame.
    \item Alignment is strictly one-to-one: \texttt{image[i]} always pairs with \texttt{action[i]}.
    \item An action entry may describe a single action or a composite action (e.g., ``turn left and move forward''), but it still corresponds to the visual state in the paired image.
\end{itemize}

\textbf{--- Core reasoning rule ---}\\[-6pt]
\begin{itemize}
    \item Always rely primarily on \textbf{visual observations} when determining how to move.
    \item Use actions only as fallback when the image is unclear.
    \item Maintain consistent spatial logic: if an object is on the left, turning left should bring it to the center view.
\end{itemize}

\textbf{--- Language and Output Style ---}\\[-6pt]
\begin{itemize}
    \item Avoid first-person narration; use \textbf{third-person, objective} instructions such as ``A sofa is on the right; turn right to face it.''
    \item Avoid narrative openings (e.g., ``The journey begins...'').
    \item Use direct commands: ``Turn right into the hallway.'', ``Walk straight past the sofa.''.
    \item Always include all necessary turning/movement instructions.
    \item Mention only key orientation-relevant landmarks (sofa, table, doorway, window).
\end{itemize}

\textbf{--- Responsibilities ---}\\[-6pt]
\begin{enumerate}
    \item[\textbf{0.}] \textbf{Trajectory summarization:}
    \begin{itemize}
        \item Summarize overall motion, room types, and representative objects.
        \item Briefly describe the starting location.
        \item Provide a concise step-by-step movement description consistent with images.
        \item End with a clear final position description.
    \end{itemize}

    \item[\textbf{1.}] \textbf{Per-step reasoning:}
    \begin{itemize}
        \item Think in first-person as the agent (camera aligned with orientation).
        \item Base reasoning on \textbf{visible evidence} in the current frame.
        \item Mention only representative, orientation-relevant objects.
        \item Use diverse spatial expressions: ``to the right'', ``just ahead'', ``past the table'', etc.
        \item Ensure geometric consistency between viewpoint and actions.
        \item If actions conflict with geometry, trust the image.
    \end{itemize}
\end{enumerate}

\textbf{--- Action rules ---}\\[-6pt]
\begin{itemize}
    \item Actions may be single or composite (joined by ``and'').
    \item Allowed actions: \texttt{TurnLeft}, \texttt{TurnRight}, \texttt{MoveForward}, \texttt{Move}, \texttt{Stop}.
    \item ``Move'' alone means a small forward motion without rotation.
    \item Composite actions operate sequentially: turn first, then move.
\end{itemize}

\textbf{--- Special Requirement: Three Reformulations ---}\\
Rewrite the trajectory summary into \textbf{three distinct linguistic styles} with identical semantic content: Formal Instructional Style, Natural Conversational Style and Narrative Descriptive Style. 

Guidelines:
\begin{itemize}
    \item All three must preserve identical spatial logic and landmarks.
    \item No conflicts are allowed.
    \item All must fully cover the entire trajectory.
\end{itemize}

\textbf{--- Examples of the Three Styles ---}
\begin{itemize}
    \item \textbf{Instruction 1 (Formal):} ``Turn right into the hallway. Advance straight past the dining table. Enter the bedroom and stop in front of the bed.''
    \item \textbf{Instruction 2 (Conversational):} ``Take a right into the hallway and keep walking until you pass the dining table on your left. Go into the bedroom and stop by the bed.''
    \item \textbf{Instruction 3 (Narrative):} ``Turning right, you move into the hallway, the dining table sliding by on your left. The hall opens into a bedroom, where you halt just before the bed.''
\end{itemize}\\

\bottomrule
\end{tabular}}
\end{table*}

\begin{table*}[t]
\centering
\caption{\textbf{Prompts to generate route plan VQA in \dataset - part1.}}
\label{tab:rp_vqa_prompt_1}
\small
\begin{tabular}{p{0.95\linewidth}}
\toprule
You are an AI assistant tasked with generating \textbf{Fill-in-the-blank Action Completion MCQ} for robot navigation. Your job is to output a multiple-choice question (with blanks) and its correct answer.

\textbf{--- Input ---}\\
A sequence of continuous key frames from a room-tour video (the frames are consecutive and represent a smooth camera/robot trajectory).

\textbf{--- High-level rule (priority order) ---}
\begin{enumerate}
\item ALL reasoning must be grounded purely on the **visual evidence from the frames** 
\item Use visual cues such as object appearance/disappearance, relative positions, scaling, and viewpoint rotation to infer the robot’s movements and turns.
\item When describing places, objects, or targets, use detailed and specific visual anchors — not just generic room names. For example: “the blue sofa on the right,” “the black dining table ahead,” “the kitchen counter with sink,” or “the hallway with a white door at the end.”
\item If any step or turn cannot be confidently inferred from visual evidence, skip or merge it rather than guessing. Do NOT fabricate movements.
\end{enumerate}

\textbf{--- Core Procedure (must follow this order) ---}

\begin{enumerate}
\item \textbf{Construct a concise, numbered Trajectory Summary}:
    \begin{itemize}
    \item Carefully analyze the continuous frame sequence to extract a minimal yet complete trajectory.
    \item Each entry in the summary should be a single action step, such as: "1. Go forward until [object/room]", "2. Turn left", "3. Go forward until [object/room]".
    \item Determine steps by tracking:
        \begin{itemize}
        \item Appearance/disappearance or scaling of landmarks (for “Go forward”)
        \item Change in viewing direction or lateral movement (for “Turn left/right/back”)
        \end{itemize}
    \item The summary should form a coherent navigation sequence from the starting viewpoint to the final destination.
    \item Explicitly describe both start place and end place in visually grounded detail: e.g., “You are a robot beginning at the living room, facing the blue sofa.” e.g., “You want to navigate to the kitchen with a table on your left.”
    \item When describing each “Go forward” anchor, be as specific as visually supported:
        \begin{itemize}
        \item Include object appearance (color, size, material), or scene context (e.g., furniture type or nearby area). Example: “Go forward until the blue sofa.”
        \end{itemize}
    \item Only include meaningful transitions — skip redundant minor movements or rotations that don’t correspond to clear spatial change.
    \item Ensure geometric reasoning consistency.
    \item Make the trajectory alternate logically between “Go forward” and “Turn”.
    \item Example of a trajectory summary:
        \begin{itemize}
        \item 1. Go forward until the 3-seater sofa (evidence: frames X–Y)
        \item 2. Turn right (evidence: frames X–Y)
        \item 3. Go forward until the dining table (evidence: frames X–Y)
        \item 4. Turn left (evidence: frames X–Y)
        \item 5. Go forward until the kitchen counter with sink (evidence: frames X–Y)
        \end{itemize}
    \end{itemize}

\item \textbf{QA Generation (based on Trajectory Summary only)}:
    \begin{itemize}
    \item Normalize steps to alternate between “Go forward…” and “Turn …”.
    \item Determine where to place \texttt{[please fill in]} blanks:
        \begin{itemize}
        \item Every turn step must become a blank.
        \end{itemize}
    \item “Go forward” must mention detailed visible landmarks.
    \item Use the strict template:
\begin{verbatim}
Q: You are a robot beginning at [start place, with visual details]. 
You want to navigate to [end place, with visual details].
You will perform the following actions:
1. Go forward until [object/room]
2. [please fill in]
3. Go forward until [object/room]
...
N. Go forward until [object/room].
You have reached the final destination.
(Note: for each [please fill in], choose either 
'turn back,' 'turn left,' or 'turn right.')
\end{verbatim}
    \end{itemize}
\end{enumerate}
\\
\bottomrule
\end{tabular}
\end{table*}

\begin{table*}[t]
\centering
\caption{\textbf{Prompts to generate route plan VQA in \dataset - part2}}
\label{tab:rp_vqa_prompt_2}
\small
\begin{tabular}{p{0.95\linewidth}}
\toprule

\textbf{3. Options generation:}
\begin{itemize}[left=2em] 
    \item For each blank, permissible options:  
        \begin{itemize}[left=2em]
        \item 'turn back', 'turn left', 'turn right'
        \end{itemize}
    \item If one blank: produce A–C.  
    \item If >= 2 blanks: produce A–D.
    \item Each option is a full sequence of turns.
\end{itemize}

\textbf{4. Correct answer determination:}
\begin{itemize}[left=2em]
    \item Must match turns inferred from visual trajectory summary.
    \item No guessing ambiguous turns.
\end{itemize}

\textbf{--- Output format (STRICT JSON SCHEMA — all keys required) ---}
\begin{itemize}[left=2em]
    \item "trajectory summary": a list of strings.
    \item "question": full question string.
    \item "options": dictionary with keys A, B, C, D.
    \item "answer": one correct option.
\end{itemize}

\textbf{--- Strict behaviour notes (must obey) ---}
\begin{itemize}[left=2em]
    \item Only use frame-based evidence.
    \item Build a reliable Trajectory Summary.
    \item The first step may be a Turn or a Go-forward action.
    \item Never guess ambiguous turns.
\end{itemize}

\\
\bottomrule
\end{tabular}
\end{table*}


\end{document}